\documentclass[11pt,a4paper]{article}
\usepackage[hyperref]{acl2019}
\usepackage{times}
\usepackage{latexsym}
\usepackage{url}
\usepackage{booktabs}
\usepackage{scrextend}
\usepackage{graphicx}
\usepackage{textpos}
\usepackage{xcolor}
\usepackage{tikz}
\usepackage[ruled]{algorithm}
\usepackage[noend]{algpseudocode}
\usepackage[normalem]{ulem}
\usepackage{colortbl}
\usepackage{microtype}
\usepackage{float}
\usepackage{placeins}

\usepackage{wasysym}
\usepackage{xcolor}

\newcommand{\RMSE}{\textsc{rmse}}
\newcommand{\QWK}{$\mathrm{qw}\kappa$}
\newcommand{\DEF}
{\mathtt{DEF}}
\newcommand{\SEL}
{\mathtt{SEL}}
\newcommand{\SIZE}{\mathtt{SIZE}}

\aclfinalcopy 
\def\aclpaperid{345} 

\title{Manipulating the Difficulty of C-Tests } 

\author{Ji-Ung Lee \and Erik Schwan \and Christian M. Meyer \\
  Ubiquitous Knowledge Processing (UKP)  Lab and Research Training Group AIPHES  \\
Computer Science Department, Technische Universit\"at Darmstadt, Germany \\
  {\url{https://www.ukp.tu-darmstadt.de}}}

\date{}

\begin{document}
\maketitle
\begin{abstract}
We propose two novel manipulation strategies for increasing and decreasing the difficulty of C-tests automatically. 
This is a crucial step towards generating learner-adaptive exercises for self-directed language learning and preparing language assessment tests.
To reach the desired difficulty level, we manipulate the size and the distribution of gaps based on absolute and relative gap difficulty predictions.
We evaluate our approach in corpus-based experiments and in a user study with 60 participants. 
We find that both strategies are able to generate C-tests with the desired difficulty level. 
\end{abstract}


\section{Introduction}
\label{sec:intro}

Learning languages is of utmost importance in an international society and formulated as a major political goal by institutions such as the European Council, who called for action to ``teaching at least two foreign languages'' \cite[p.\ 20]{EC02}.
But also beyond Europe, there is a huge demand for language learning worldwide 
due to increasing globalization, digital communication, and migration.

Among multiple different learning activities required for effective language learning, we study one particular type of exercise in this paper:
\emph{C-tests} are a special type of cloze test in which the second half of every second word in a given text is replaced by a gap \cite{Klein1982}.
Figure~\ref{fig:examples} (a) shows an example.
To provide context, the first and last sentences of the text do not contain any gaps.
C-tests rely on the reduced redundancy principle \cite{Spolsky69} arguing that a language typically employs more linguistic information than theoretically necessary to communicate unambiguously.
Proficient speakers intuitively understand an utterance even if the level of redundancy is reduced (e.g., when replacing a word's suffix with a gap), whereas learners typically rely on the redundant signal to extrapolate the meaning of an utterance.

Besides general vocabulary knowledge, C-tests require orthographic, morphologic, syntactic, and semantic competencies \cite{Chapelle94} to correctly fill in all gaps, which make them a frequently used tool for language assessment (e.g., placement tests).
Given that C-tests can be easily generated automatically by introducing gaps into an arbitrary text and that there is usually only a single correct answer per gap given its context, C-tests are also relevant for self-directed language learning and massive open online courses (MOOC), where large-scale personalized exercise generation is necessary.

A crucial question for such tasks is predicting and manipulating the \emph{difficulty} of a C-test.
For language assessment, it is important to generate C-tests with a certain target difficulty to allow for comparison across multiple assessments.
For self-directed language learning and MOOCs, it is important to adapt the difficulty to the learner's current skill level, as an exercise should be neither too easy nor too hard so as to maximize the learning effect and avoid boredom and frustration \cite{Vygotsky78}.
Automatic difficulty prediction of C-tests is hard, even for humans, which is why there have been many attempts to theoretically explain C-test difficulty (e.g., \citealp{Sigott1995}) and to model features used in machine learning systems for automatic difficulty prediction (e.g., \citealp{Beinborn2014}).

\begin{figure*}
	\begingroup
	\small
	\arrayrulecolor{gray}
	\newcolumntype{L}[1]{>{\ttfamily\small\raggedright\arraybackslash}m{#1}}
	\begin{tabular}{|*{3}{@{ }L{.32\textwidth}@{ }|}}
		\hline
		It i\_ being fou\_\_\_, moreover, i\_ fairly cl\_\_\_ correspondence wi\_\_ the predi\_\_\_\_\_\_ of t\_\_ soothsayers o\_ the th\_\_\_ factories. Th\_\_ predicted escal\_\_\_\_\_, and escal\_\_\_\_\_ is wh\_\_ we a\_\_ getting. T\_\_ biggest nuc\_\_\_\_ device t\_\_ United Sta\_\_\_ has expl\_\_\_\_ measured so\_\_ 15 meg\dots
		& 
		It is being fought, more\_\_\_\_, in fai\_\_\_ cl\_\_\_ corresp\_\_\_\_\_\_\_ with the predi\_\_\_\_\_\_ of the sooth\_\_\_\_\_\_ of the th\_\_\_ fact\_\_\_\_\_. Th\_\_ pred\_\_\_\_\_ escal\_\_\_\_\_, and escal\_\_\_\_\_ is what w\_ are get\_\_\_\_. The big\_\_\_\_ nuc\_\_\_\_ dev\_\_\_ the United States h\_\_ expl\_\_\_\_ meas\_\_\_\_ some 15 meg\dots
		& 
		It i\_ being fough\_, moreover, i\_ fairly clos\_ correspondence wit\_ the prediction\_ of t\_\_ soothsayers o\_ the thin\_ factories. The\_ predicted escalatio\_, and escalatio\_ is wha\_ we ar\_ getting. T\_\_ biggest nuclea\_ device t\_\_ United State\_ has explode\_ measured som\_ 15 meg\dots
		\\
		\hline
		\multicolumn{1}{c}{(a)} & 
		\multicolumn{1}{c}{(b)} & 
		\multicolumn{1}{c}{(c)} \\[-3pt]
	\end{tabular}
	\endgroup
	\caption{C-tests with (a) standard gap scheme, (b) manipulated gap position, and (c) manipulated gap size}
	\label{fig:examples}
\end{figure*}

While state-of-the-art systems produce good prediction results compared to humans \cite{Beinborn2016}, there is yet no work on \emph{automatically manipulating} the difficulty of C-tests.
Instead, C-tests are generated according to a fixed scheme and manually post-edited by teachers, who might use the predictions as guidance.
But this procedure is extremely time-consuming for language assessment and no option for large-scale self-directed learning.

In this paper, we propose and evaluate two strategies for automatically changing the gaps of a C-test in order to reach a given target difficulty.
Our first strategy varies the distribution of the gaps in the underlying text and our second strategy learns to decide to increase or decrease a gap in order to make the test easier or more difficult.
Our approach breaks away from the previously fixed C-test creation scheme and explores new ways of motivating learners by using texts they are interested in and generating tests from them at the appropriate level of difficulty.
We evaluate our strategies both automatically and in a user study with 60 participants.

\section{Related Work}
\label{sec:related}

In language learning research, there is vast literature on cloze tests.
For example, \citet{Taylor1953} studies the relation of cloze tests and readability.
In contrast to C-tests \citep{Klein1982}, cloze tests remove whole words to produce a gap leading to more ambiguous solutions.

\citet{Chapelle90} contrast four types of cloze tests, including fixed-ratio cloze tests replacing every $i$\textsuperscript{th} word with a gap, rational cloze tests that allow selecting the words to replace according to the language trait that should be assessed, multiple-choice tests, and C-tests.
Similar to our work, they conduct a user study and measure the difficulty posed by the four test types.
They find that cloze tests replacing entire words with a gap are more difficult than C-tests or multiple-choice tests.
In our work, we go beyond this by not only varying between gaps spanning the entire word (cloze test) or half of the word (C-test), but also changing the size of the C-test gaps.
\Citet{Laufer99} propose using C-tests to assess vocabulary knowledge.
To this end, they manually construct C-tests with only a single gap, but use larger gaps than half of the word's letters.
Our work is different to these previous works, since we test varying positions and sizes for C-test gaps and, more importantly, we aim at manipulating the difficulty of a C-test automatically by learning to predict the difficulty of the gaps and how their manipulation affects the difficulty.

Previous work on automatically controlling and manipulating test difficulty has largely focused on multiple-choice tests by generating appropriate distractors (i.e., incorrect solutions).
\Citet{Wojatzki2016} avoid ambiguity of their generated distractors, \citet{Hill2016} fit them to the context, and \citet{Perez2017} consider multiple languages.
Further work by \citet{Zesch2014}, \citet{Beinborn2016}, and \citet{Lee2016} employ word difficulty, lexical substitution, and the learner's answer history to control distractor difficulty.

For C-tests, \citet{Kamimoto93} and \citet{Sigott2006} study  features of hand-crafted tests that influence the difficulty, and \citet{Beinborn2014} and \citet{Beinborn2016} propose an automatic approach to estimate C-test difficulty, which we use as a starting point for our work.

Another related field of research in computer-assisted language learning is readability assessment and, subsequently, text simplification.
There exists ample research on predicting the reading difficulty for various learner groups \cite{Hancke2012,Collins2014,Pilan2014}.
A specific line of research focuses on reducing the reading difficulty by text simplification \cite{Chandrasekar1996}.
By reducing complex texts or sentences to simpler ones, more texts are made accessible for less proficient learners.
This is done either on a word level by substituting difficult words with easier ones (e.g., \citealp{Kilgarriff2014}) or on a sentence level \cite{Vajjala2014}.
More recent work also explores sequence-to-sequence neural network architectures for this task \cite{Nisioi2017}.
Although the reading difficulty of a text partly contributes to the overall exercise difficulty of C-tests, there are many other factors with a substantial influence \cite{Sigott1995}.
In particular, we can generate many different C-tests from the same text and thus reading difficulty and text simplification alone are not sufficient to determine and manipulate the difficulty of C-tests.

\section{Task Overview}
\label{sec:task}

We define a C-test $T = (u, w_1,\dots, w_{2n}, v, G)$ as a tuple of left and right context $u$ and $v$ (typically one sentence) enframing $2n$ words $w_i$ where $n\!=\!|G|$ is the number of gaps in the gap set $G$.
In each gap $g\!=\!(i,\!\ell)\!\in G$, the last $\ell$ characters of word $w_i$ are replaced by a blank for the learners to fill in.
\Citet{Klein1982} propose the default gap generation scheme $\DEF$ with $G = \{(2j, \lceil\frac{|w_{2j}|}{2}\rceil) \mid 1\!\le\!j\!\le\!n\}$ in order to trim the (larger) second half of every second word.
Single-letter words, numerals, and punctuation are not counted as words $w_i$ and thus never contain gaps.
Figure~\ref{fig:examples} (a) shows an example C-test generated with the $\DEF$ scheme.

A major limitation of $\DEF$ is that the difficulty of a C-test is solely determined by the input text.
Most texts, however, yield a medium difficulty (cf.\ section~\ref{sec:analysis}) and thus do not allow any adaptation to beginners or advanced learners unless they are manually postprocessed.
In this paper, we therefore propose two strategies to manipulate the gap set $G$ in order to achieve a given \emph{target difficulty} $\tau \in [0,1]$ ranging from small values for beginners to high values for advanced learners. 
To estimate the difficulty $d(T) = \frac{1}{|G|} \sum_{g \in G} d(g)$ of a C-test $T$, we aggregate the predicted difficulty scores $d(g)$ of each gap.
In section~\ref{sec:prediction}, we reproduce the system by \citet{Beinborn2016} modeling $d(g)\approx e(g)$ as the estimated mean error rates $e(g)$ per gap across multiple learners, and we conduct additional validation experiments on a newly acquired dataset.

\begin{figure}
	\centering
	\includegraphics[width=.98\linewidth,trim=0cm 2.05cm 0cm 0cm,clip]{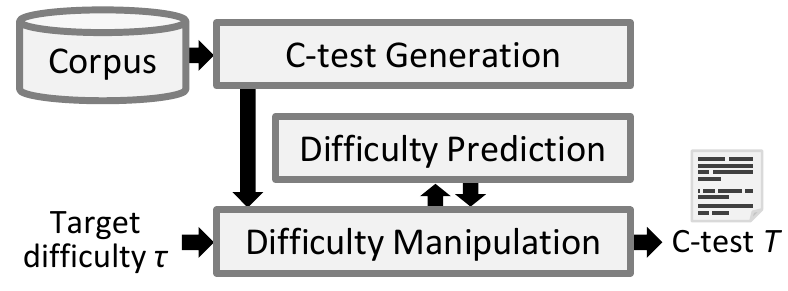}
	\caption{Proposed system architecture}
	\label{fig:overview}
\end{figure}

The core of our work is the manipulation of the gap set $G$ in order to minimize the difference $|d(T)-\tau|$ between the predicted test difficulty $d(T)$ and the requested target difficulty $\tau$.
To this end, we employ our difficulty prediction system for  validation and propose a new regression setup that predicts the relative change of $d(g)$ when manipulating the size $\ell$ of a gap.

Figure~\ref{fig:overview} shows our system architecture: Based on a text corpus, we generate C-tests for arbitrary texts (e.g., according to the learner's interests).
Then, we manipulate the difficulty of the generated text by employing the difficulty prediction system in order to reach the given target difficulty $\tau$ for a learner (i.e., the estimated learner proficiency) to provide neither too easy nor too hard tests.

\section{C-Test Difficulty Prediction}
\label{sec:prediction}

\Citet{Beinborn2014} and \citet{Beinborn2016} report state-of-the-art results for the C-test difficulty prediction task.
However, there is yet no open-source implementation of their code and there is little knowledge about the performance of newer approaches.
Therefore, we (1)~conduct a reproduction study of \citeauthor{Beinborn2016}'s (\citeyear{Beinborn2016}) system, (2)~evaluate newer neural network architectures, and (3)~validate the results on a newly acquired dataset.

\paragraph{Reproduction study.}
We obtain the original software and data from \citet{Beinborn2016}.
This system predicts the difficulty $d(g)$ for each gap within a C-test using a support vector machine (SVM; \citealp{Vapnik1998}) with 59 hand-crafted features.
The proposed features are motivated by four factors which are deemed important for assessing the gap difficulty: \textit{item dependency}, \textit{candidate ambiguity}, \textit{word difficulty}, and \textit{text difficulty}. 
We use the same data (819 filled C-tests), metrics, and setup as \citet{Beinborn2016}.
That is, we perform leave-one-out cross validation (LOOCV) and measure the Pearson correlation $\rho$, the rooted mean squared error \RMSE{}, and the quadratic weighted kappa \QWK{} as reported in the original work.

The left hand side of table~\ref{tab:reproduction-experiments} shows the results of our reproduced SVM 
compared to the original SVM results reported by \citet{Beinborn2016}.
Even though we reuse the same code as in their original work, we observe small differences between our reproduction and the previously reported scores. 

We were able to trace these differences back to libraries and resources which have been updated and thus changed over time.
One example is Ubuntu's system dictionary, the \textit{American English dictionary words} (wamerican), on which the original system relies. 
We experiment with different versions of the dictionary between Ubuntu 14.04 (wamerican v.7.1.1) and 18.04 (wamerican v.2018.04.16-1) and observe differences of one or two percentage points.
As a best practice, we suggest to fix the versions of all resources and avoid any system dependencies.

\begin{table}
	\centering\small
	\begin{tabular}{l*{2}{@{\hspace{.3cm}}*{3}{@{\hspace{.15cm}}c}}}
		\toprule
		& \multicolumn{3}{c}{\textbf{Original data}} 
		& \multicolumn{3}{c}{\textbf{New data}} \\
		Model & $\rho$ & \RMSE{} & \QWK & $\rho$ & \RMSE{} & \QWK \\
		\midrule
		SVM (original)   & .50 & .23 & .44 & -- & -- & -- \\
		SVM (reproduced) & .49 & .24 & .47 & .50 & .21 & .39 \\
		MLP              & .42 & .25 & .31 & .41 & .22 & .25 \\
		BiLSTM           & .49 & .24 & .35 & .39 & .24 & .27 \\
		\bottomrule
	\end{tabular}
	\caption{Results of the difficulty prediction approaches. SVM (original) has been taken from \citet{Beinborn2016}}\label{tab:reproduction-experiments}
\end{table}

\paragraph{Neural architectures.}
We compare the system with deep learning methods based on multi-layer perceptrons (MLP) and bi-directional long short-term memory (BiLSTM) architectures, which are able to capture non-linear feature dependencies.\footnote{Network parameters and a description of the tuning process are provided in this paper's appendix.}
To cope for the non-deterministic behavior of the neural networks, we repeat all experiments ten times with different random weight initializations and report the averaged results \cite{Reimers2017}.
While the MLP is trained similar as our reproduced SVM, the BiLSTM receives all gaps of a C-test as sequential input. 
We hypothesize that this sequence regression setup is better suited to capture gaps interdependencies.
As can be seen from the table, the results of the neural architectures are, however, consistently worse than the SVM results.
We analyze the \RMSE{} on the train and development sets and observe a low bias, but a high variance.
Thus, we conclude that although neural architectures are able to perform well for this task, they lack a sufficient amount of data to generalize.

\paragraph{Experiments on new data.}
To validate the results and assess the robustness of the difficulty prediction system, we have acquired a new C-test dataset from our university's language center.
803 participants of placement tests for English courses solved five C-tests (from a pool of 53 different C-tests) with 20 gaps each.
Similar to the data used by \citet{Beinborn2016}, we use the error rates $e(g)$ for each gap as the $d(g)$ the methods should predict.

The right-hand side of table~\ref{tab:reproduction-experiments} shows the performance of our SVM and the two neural methods.
The results indicate that the SVM setup is well-suited for the difficulty prediction task and that it successfully generalizes to new data.

\paragraph{Final model.}
We train our final SVM model on all available data (i.e.,\ the original and the new data)
and publish our source code and the trained model on GitHub.\footnote{\url{https://github.com/UKPLab/acl2019-ctest-difficulty-manipulation} \linebreak(licensed under the Apache License 2.0).}
Similar to \citet{Beinborn2016}, we cannot openly publish our dataset due to copyright.

\section{C-Test Difficulty Manipulation}
\label{sec:manipulation}

\begin{algorithm}[t]
	\small
	\begin{algorithmic}[1]
		\newcommand{\Gfull}{G_\mathtt{FULL}}
		\algrenewcommand\algorithmicindent{1.25em}
		\Procedure{GapSelection}{$T$, $\tau$}
		\State $\Gfull \gets \{(i, \lceil\frac{|w_i|}{2}\rceil \mid 1 \le i \le 2n\}$
		\State $G_\SEL \gets \emptyset$
		\While {$|G_\SEL| < n$}
		\State $G_{\le \tau} \gets \{g \in \Gfull \mid d(g) \le \tau\}$
		\If {$|G_{\le \tau}| > 0$}
		\State $g^* \gets \arg\min_{g\in G_{\le \tau}} |d(g) - \tau|$
		\State $G_\SEL \gets G_\SEL \cup \{g^*\} $
		\State $\Gfull \gets \Gfull \setminus \{g^*\}$
		\EndIf
		\State $G_{> \tau} \gets \{g \in \Gfull \mid d(g) > \tau\}$
		\If {$|G_{> \tau}| > 0$}
		\State $g^* \gets \arg\min_{g\in G_{> \tau}} |d(g) - \tau|$
		\State $G_\SEL \gets G_\SEL \cup \{g^*\} $
		\State $\Gfull \gets \Gfull \setminus \{g^*\}$
		\EndIf
		\EndWhile
		\vspace*{-.1cm}
		\State \Return $G_\SEL$     
		\EndProcedure
	\end{algorithmic}
	\caption{Gap selection strategy ($\SEL$)}
	\label{alg:selection}
\end{algorithm}

Given a C-test $T\!=\!(u, w_1,\ldots, w_{2n}, v, G)$ and a target difficulty $\tau$, the goal of our manipulation strategies is to find a gap set $G$ such that $d(T)$ approximates $\tau$.
A na\"ive way to achieve this goal would be to generate C-tests for all texts in a large corpus with the $\DEF$ scheme and use the one with minimal $|d(T) - \tau|$.
However, most corpora tend to yield texts of a limited difficulty range that only suit a specific learner profile (cf.\ section~\ref{sec:analysis}).
Another drawback of the na\"ive strategy is that it is difficult to control for the topic of the underlying text and in the worst case, the necessity to search through a whole corpus for selecting a fitting C-test.

In contrast to the na\"ive strategy, our proposed manipulation strategies are designed to be used in real time and manipulate any given C-test within 15 seconds at an acceptable quality.\footnote{On an \textit{Intel-i5} with 4 CPUs and 16 GB RAM.}
Both strategies operate on a given text (e.g., on a topic a learner is interested in) and manipulate its gap set $G$ in order to come close to the learner's current language skill.
The first strategy varies the position of the gaps and the second strategy learns to increase or decrease the size of the gaps.

\subsection{Gap Selection Strategy}
The default C-test generation scheme $\DEF$ creates a gap in every second word $w_{2j}$, $1\!\le\!j\!\le\!n$.
The core idea of our first manipulation strategy $\SEL$ is to distribute the $n$ gaps differently among the all $2n$ words in order to create gaps for easier or harder words than in the default generation scheme.
Therefore, we use the difficulty prediction system to predict $d(g)$ for any possible 
gap $g \in G_\mathtt{FULL} = \{(i, \lceil\frac{|w_i|}{2}\rceil) \mid 1 \le i \le 2n\}$ (i.e., assuming a gap in all words rather than in every second word). 
Then, we alternate between adding gaps to the resulting $G_\SEL$ that are easier and harder than the preferred target difficulty $\tau$, starting with those having a minimal difference $|d(g) - \tau|$.

Algorithm~\ref{alg:selection} shows this procedure in pseudocode and figure~\ref{fig:examples} shows a C-test whose difficulty has been increased with this strategy.  Note that it has selected gaps at \textit{corresponding} rather than \textit{with}, and \textit{soothsayers} rather than \textit{the}.
Our proposed algorithm is optimized for runtime.
An exhaustive search would require testing ${2n \choose n}$ combinations if the number of gaps is constant. For $n\!=\!20$, this yields 137 billion combinations.
While more advanced optimization methods might find better gap selections, we show in section~\ref{sec:analysis} that our strategy achieves good results.

\subsection{Gap Size Strategy}

Our second manipulation strategy $\SIZE$ changes the size of the gaps based on a pre-defined gap set. 
Increasing a gap $g\!=\!(i, \ell)$ by one or more characters, yielding $g'\!=\!(i, \ell + k)$ increases its difficulty (i.e., $d(g')\!\ge\!d(g)$), while smaller gaps make the gap easier.
We identify a major challenge in estimating the effect of increasing or decreasing the gap size on the gap difficulty.
Although $d(g')$ could be estimated using the full difficulty prediction system, the search space is even larger than for the gap selection strategy, since each of the $n$ gaps has $|w_i|\!-\!2$ possible gap sizes to test. For $n =20$ and an average word length of six, this amounts to one trillion possible combinations.

We therefore propose a new approach to predict the \emph{relative difficulty change} of a gap $g\!=\!(i, \ell)$ when increasing the gap size by one letter $\Delta_\mathrm{inc}(g) \approx d(g') - d(g)$, $g' = (i, \ell + 1)$ and correspondingly when decreasing the gap size by one letter $\Delta_\mathrm{dec}(g) \approx d(g) - d(g')$, $g' = (i, \ell - 1)$.
The notion of relative difficulty change 
enables gap size manipulation in real time, since we do not have to invoke the full difficulty prediction system for all combinations. Instead, we can incrementally predict the effect of changing a single gap.

To predict $\Delta_\mathrm{inc}$ and $\Delta_\mathrm{dec}$, we train two SVMs on all gap size combinations of 120 random texts from the Brown corpus \cite{Francis1965} using the following features: 
predicted absolute gap difficulty, word length, new gap size, modified character, a binary indicator if the gap is at a \textit{th} sound, and logarithmic difference of alternative solutions capturing the degree of ambiguity with varying gap size.

With a final set of only six features, our new models are able to approximate the relative difficulty change very well deviating from the original system's prediction only by $0.06$ \RMSE{} for $\Delta_\mathrm{inc}$ and $0.13$ \RMSE{} for $\Delta_\mathrm{dec}$.
The predictions of both models highly correlate with the predictions achieving a Pearson's $\rho$ of over $0.8$.
Besides achieving a much faster average runtime of 0.056 seconds for the relative model vs.\ 11 seconds for the full prediction of a single change, we can invoke the relative model iteratively to estimate $d(T)$ for multiple changes of the gap size more efficiently.

\begin{algorithm}[t]
	\small
	\begin{algorithmic}[1]
		\algrenewcommand\algorithmicindent{1.25em}
		\Procedure{IncreaseDifficulty}{$T$, $\tau$}
		\State $G_\SIZE \gets G_\DEF$
		\State $D \gets d(T)$
		\While {$D < \tau$}
		\State $g^* = (i, \ell) \gets \arg\max_{g \in G_\SIZE}\Delta_\mathrm{inc}(g)$
		\State $\ell \gets \ell + 1$
		\State $D \gets D + \Delta_\mathrm{inc}(g)$
		\EndWhile
		\vspace*{-.1cm}
		\State \Return $G_\SIZE$
		\EndProcedure
	\end{algorithmic}
	\caption{Gap size strategy ($\SIZE$)}
	\label{alg:size}
\end{algorithm}

The final manipulation strategy then requires just a single call of the full prediction system. 
If $d(T)\!<\!\tau$, we incrementally increase the gap sizes to make $T$ more difficult and, vice-versa, decrease the gap sizes if $d(T) > \tau$.
In each iteration, we modify the gap with the highest relative difficulty change in order to approach the given target difficulty $\tau$ as quickly as possible.
Algorithm~\ref{alg:size} shows pseudocode for creating $G_\mathrm{size}$ with increased difficulty (i.e., $d(T)\!<\!\tau$) based on the default gap scheme $\DEF$.
The procedure for $d(T)\!>\!\tau$ works analogously, but using $\Delta_\mathrm{dec}$ and decreasing the gap size.
Figure~\ref{fig:examples} (c) shows a much easier version of the example C-test, in which a learner often only has to complete the last one or two letters.

\section{Evaluation of the Manipulation System}
\label{sec:manip-eval}
\label{sec:analysis}

To evaluate our C-test manipulation strategies, we first test their ability to cover a higher range of target difficulties than the default generation scheme and then measure how well they meet the desired target difficulty for texts from different domains.
We conduct our experiments on 1,000 randomly chosen paragraphs for each of the Gutenberg  \cite{Lahiri2014}, Reuters \cite{Lewis2004}, and Brown \cite{Francis1965} corpora.
We conduct our experiments on English, but our strategies can be adapted to many related languages. 

\begin{figure}
	\centering
	\includegraphics[width=.49\textwidth,trim=.1cm .05cm 0cm .25cm,clip]{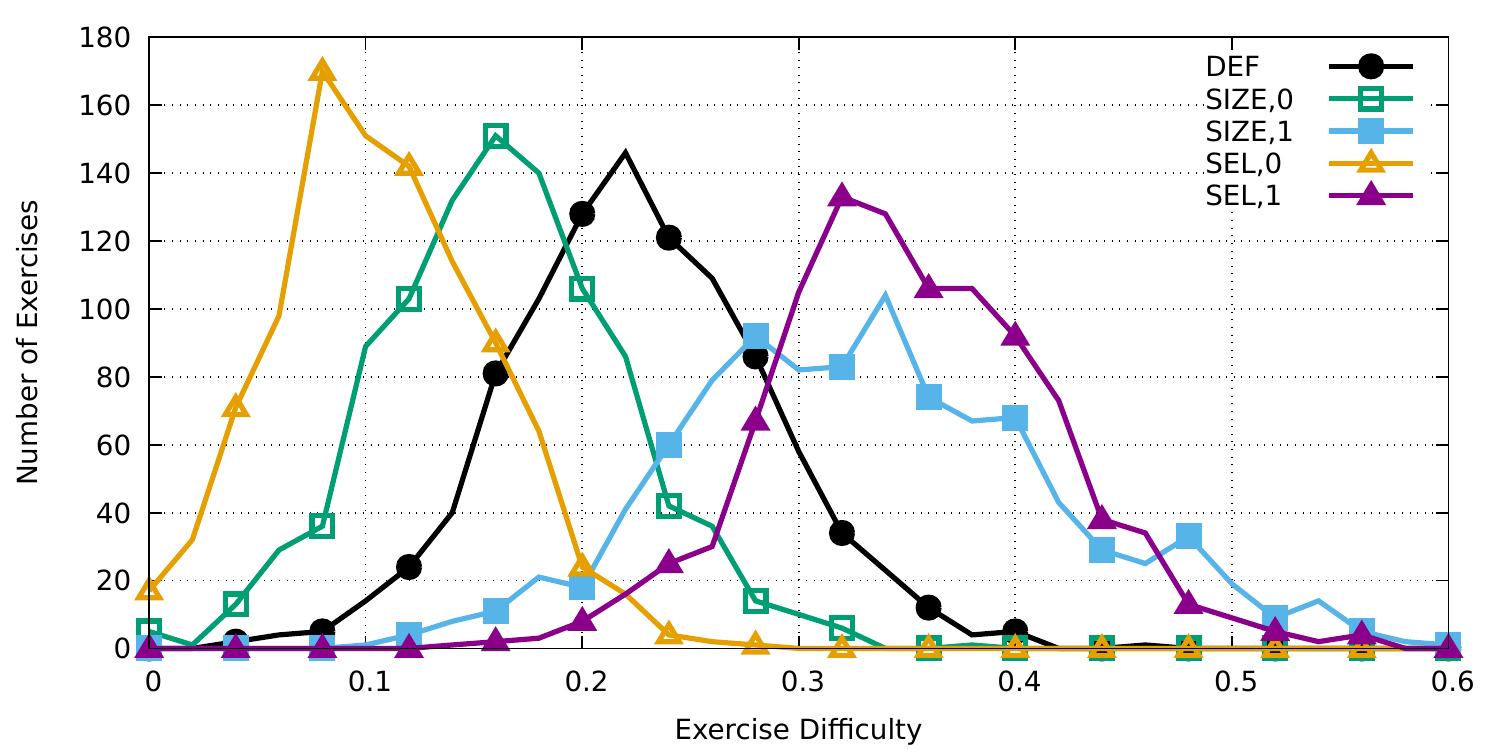}
	\caption{Difficulty distribution of exercises generated with $\DEF$, $\SEL$, and $\SIZE$ for extreme $\tau$ values}
	\label{fig:base-difficulties-brown}
\end{figure}

\paragraph{Difficulty range.}
The black $\CIRCLE$-marked line of figure~\ref{fig:base-difficulties-brown} shows the distribution of $d(T)$ based on our difficulty prediction system when creating a C-test with the default generation scheme $\DEF$ for all our samples of the Brown corpus. 
The vast majority of C-tests range between 0.15 and 0.30 with a predominant peak at 0.22.

To assess the maximal difficulty range our strategies can achieve, we generate C-tests with maximal ($\tau\!=\!1$) and minimal target difficulty ($\tau\!=\!0$) for both strategies $S\!\in\!\{\SEL, \SIZE\}$, which are also shown in figure~\ref{fig:base-difficulties-brown} as $(S, \tau)$.
Both strategies are able to clearly increase and decrease the test difficulty in the correct direction and they succeed in substantially increasing the total difficulty range beyond $\DEF$.
While $\SEL$ is able to reach lower difficulty ranges, it has bigger issues with generating very difficult tests.
This is due to its limitation to the fixed gap sizes, whereas $\SIZE$ can in some cases create large gaps that are ambiguous or even unsolvable.
Since $\SIZE$ is, however, limited to the 20 predefined gaps, 
it shows a higher variance.
Especially short gaps such as \textit{is} and \textit{it} cannot be made more difficult.
Combining the two strategies is thus a logical next step for future work, building upon our findings for both strategies.
We make similar observations on the Reuters and Gutenberg corpora and provide the respective figures in the appendix.

\paragraph{Manipulation quality.}
We finally evaluate how well each strategy $S$ reaches a given target difficulty.
That is, we sample a random corpus text and $\tau$, create the C-test using strategy $S$, predict the test difficulty $d(T)$ and measure its  difference to $\tau$ using \RMSE{}.
Table~\ref{tab:numerical-eval} shows the results for our three corpora.
Throughout all three corpora, both  manipulation strategies perform well.
$\SEL$ consistently outperforms $\SIZE$, which matches our observations from the previous experiment.
Mind that these results depend on the quality of the automatic difficulty predictions, which is why we conduct a user-based evaluation in the next section.

\begin{table}
	\centering\small
	\begin{tabular}{lccc}
		\toprule
		Strategy & Brown & Reuters & Gutenberg \\
		\midrule
		$\SEL$ & .11 & .12 & .10 \\
		$\SIZE$ & .13 & .15 & .12 \\
		\bottomrule
	\end{tabular}
	\caption{\RMSE{} for both strategies on each corpora with randomly sampled target difficulties $\tau$}\label{tab:numerical-eval}
\end{table}

\section{User-based Evaluation}
\label{sec:evaluation}
\paragraph{Hypothesis.}
To evaluate the effectiveness of our manipulation strategies in a real setting, we conduct a user study and analyze the difficulty of the manipulated and unmanipulated C-tests.
We investigate the following hypothesis:
When increasing a test's difficulty using strategy $S$, the participants will make more errors and judge the test harder than a default C-test and, vice versa, when decreasing a test's difficulty using $S$, the participants will make less errors and judge the test easier.

\paragraph{Experimental design.}
We select four different English texts from the Brown corpus and shorten them to about 100 words with keeping their paragraph structure intact.
None of the four texts is particularly easy to read with an average grade level above 12 and a Flesh reading ease score ranging between 25 (very difficult) to 56 (fairly difficult).
In the supplementary material, we provide results of an automated readability analysis using standard metrics.
From the four texts, we then generate the C-tests $T_i$, $1 \le i \le 4$ using the default generation scheme $\DEF$.
All tests contain exactly $n = 20$ gaps and their predicted difficulties $d(T_i)$ are in a mid range between 0.24 and 0.28.
$T_1$ remains unchanged in all test conditions and is used to allow the participants to familiarize with the task.
For the remaining three texts, we generate an easier variant $T_i^{S,\mathrm{dec}}$ with target difficulty $\tau = 0.1$ and a harder variant $T_i^{S, \mathrm{inc}}$ with $\tau = 0.5$ for both strategies $S \in \{\SEL, \SIZE\}$.

From these tests, we create 12 sequences of four C-tests that we give to the participants.
Each participant receives $T_1$ first to familiarize with the task.
Then, they receive one easy $T_i^{S,\mathrm{dec}}$, one default $T_i$, and one hard $T_i^{S,\mathrm{inc}}$ C-test for the same strategy $S$ based on the texts $i \in \{2,3,4\}$ in random order without duplicates (e.g., the sequence $T_1^{\vphantom{\SEL,}} \, T_2^{\SEL,\mathtt{dec}} \, T_3^{\vphantom{\SEL,}} \, T_4^{\SEL,\mathtt{inc}}$).
Having finished a C-test, we ask them to judge the difficulty of this test on a five-point Likert scale ranging from \emph{too easy} to \emph{too hard}.
After solving the last test, we additionally collect a ranking of all four tests by their difficulty.

\paragraph{Data collection.} 
We collect the data from our participants with a self-implemented web interface for solving C-tests.
We create randomized credentials linked to a unique ID for each participant and obfuscate their order, such that we can distinguish them but cannot trace back their identity and thus avoid collecting any personal information.
Additionally, we ask each participant for their consent on publishing the collected data.
For experiments with a similar setup and task, we obtained the approval of the university's ethics commission.
After login, the participants receive instructions and provide a self-assessment of their English proficiency and their time spent on language learning.
The participants then solve the four successive C-tests without knowing the test difficulty or the manipulation strategy applied.
They are instructed to spend a maximum of five minutes per C-test to avoid time-based effects and to prevent them from consulting external resources, which would bias the results.

\paragraph{Participants.}
A total of 60 participants completed the study.
We uniformly distributed the 12 test sequences (six per strategy), such that we have 30 easy, 30 default, and 30 hard C-test results for each manipulation strategy.
No participant is native in English, 17 are taking language courses, and 57 have higher education or are currently university students.
The frequency of their use of English varies, as we found a similar number of participants using English daily, weekly, monthly, and (almost) never in practice. An analysis of the questionnaire is provided in the paper's appendix.


\begin{figure*}
	\small
	\newcolumntype{L}[1]{>{\ttfamily\small\centering\arraybackslash}m{#1}}
	\begin{tabular}{*{3}{@{ }L{.32\textwidth}@{ }}}
		\includegraphics[width=0.30\textwidth]{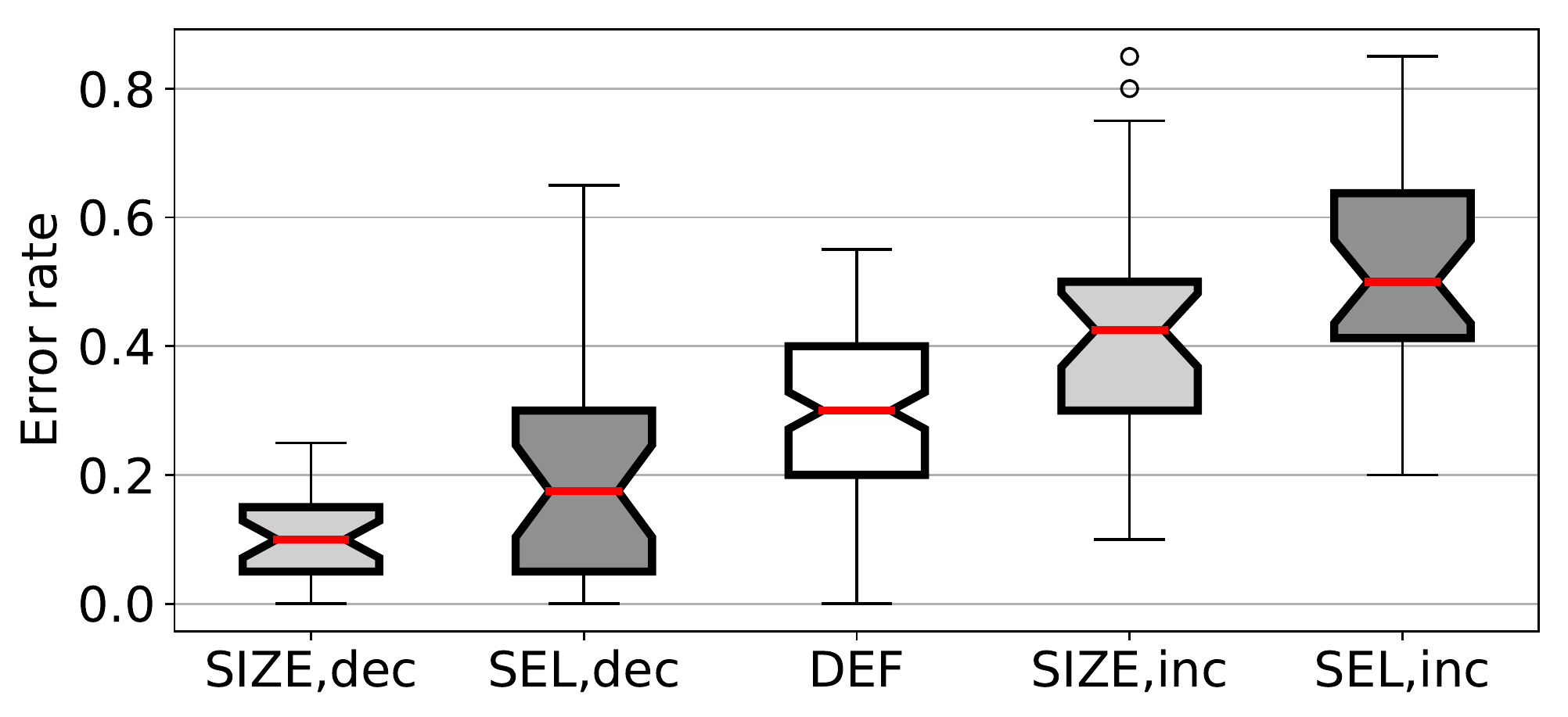}
		& 
		\includegraphics[width=0.33\textwidth]{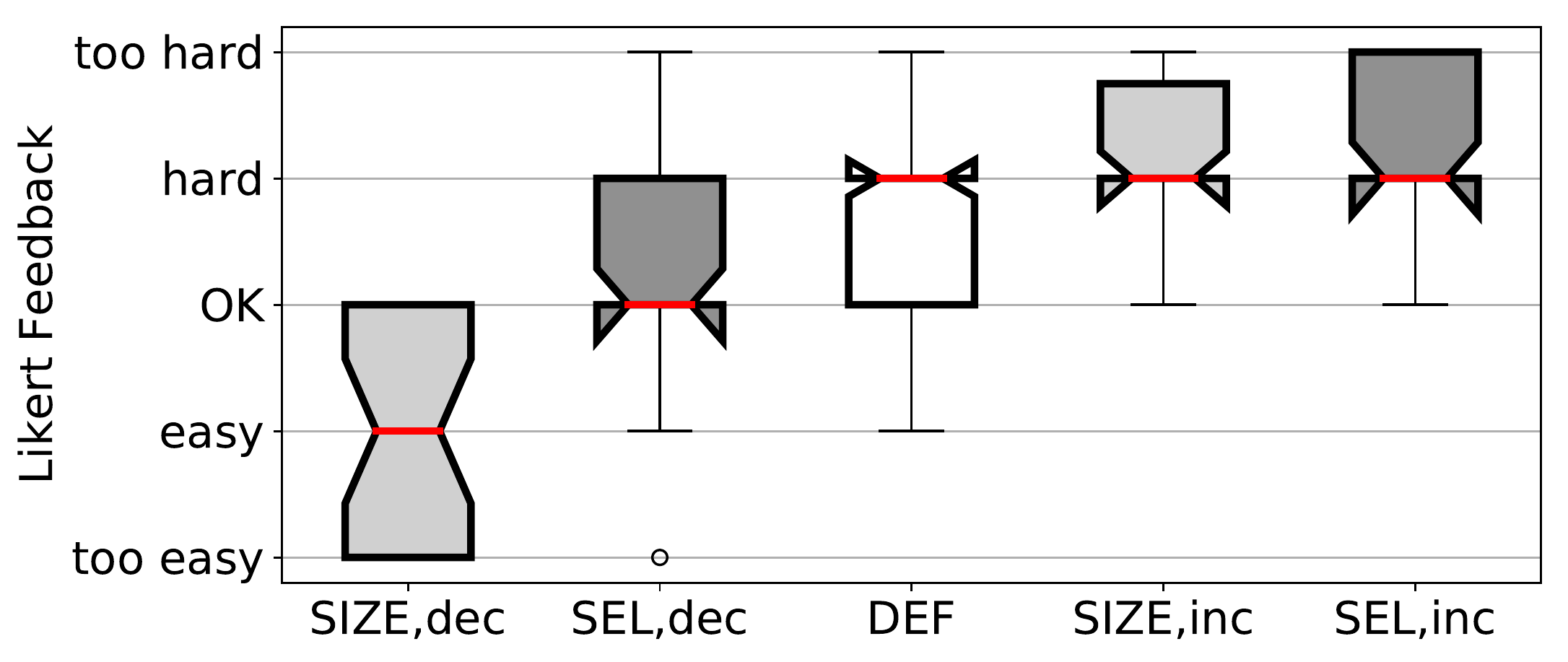}
		& 
		\includegraphics[width=0.29\textwidth]{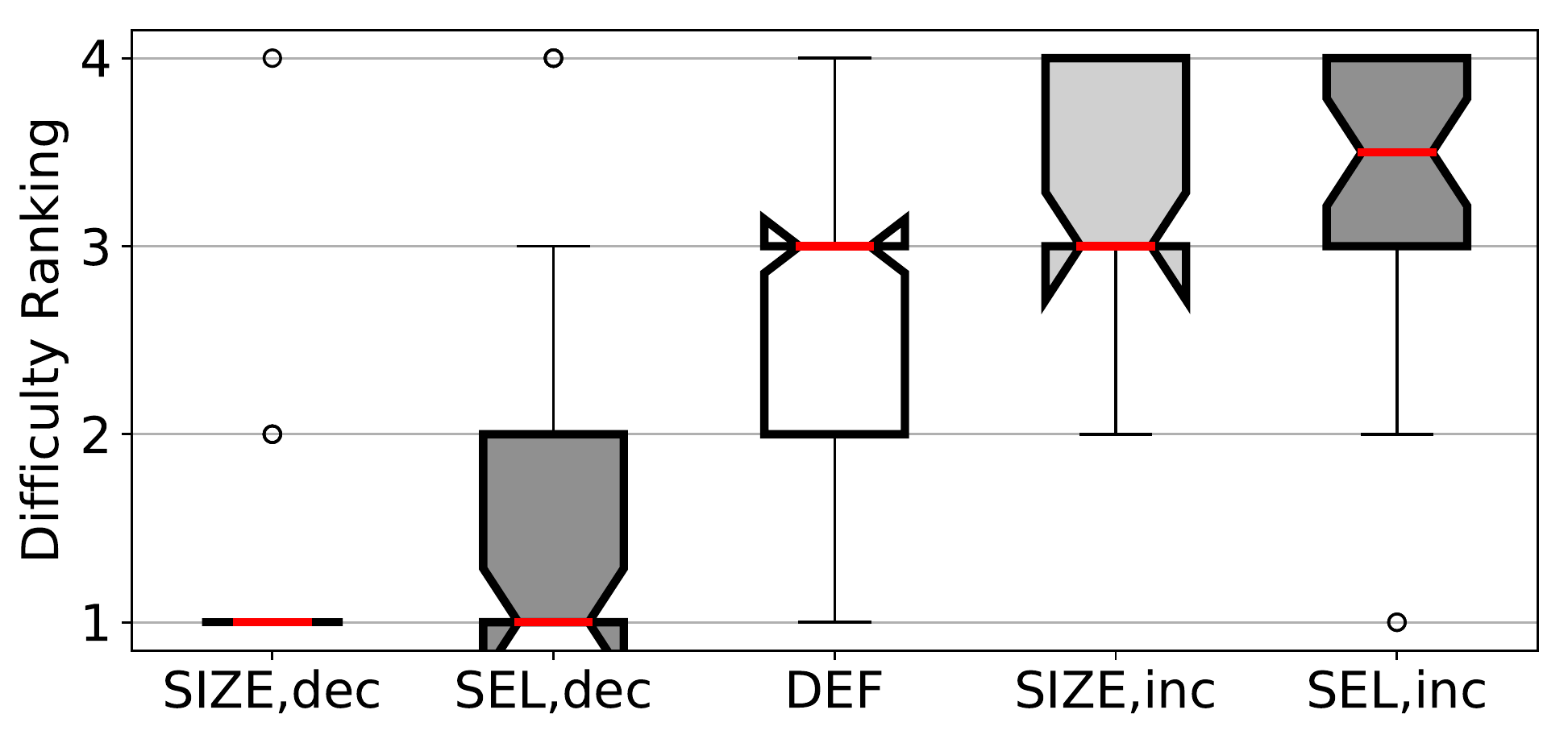}
		\\
		\multicolumn{1}{c}{(a)} & 
		\multicolumn{1}{c}{(b)} & 
		\multicolumn{1}{c}{(c)} \\[-6pt]
	\end{tabular}
	\caption{Notched boxplots for the (a) observed error rates, (b) Likert feedback, and (c) the participants' rankings}
	\label{fig:boxplots}
\end{figure*}


\paragraph{Hypothesis testing.} 
We evaluate our hypothesis along three dimensions: (1) the actual error rate of the participants, (2) the perceived difficulty after each individual C-test (Likert feedback), and (3) the participants' final difficulty ranking.
While the latter forces the participants to provide an explicit ranking, the former allows them to rate C-tests equally difficult.
We conduct significance testing at the Bonferroni-corrected $\alpha = \frac{0.05}{2} = 0.025$ for each dimension using one-tailed $t$-tests for the continuous error rates and one-tailed Mann--Whitney $U$ tests for the ordinal-scaled perceived difficulties and rankings.
Figure~\ref{fig:boxplots} shows notched boxplots of our results. 

To test our hypothesis, we first formulate a null hypothesis that (a) the mean error rate, (b) the median perceived difficulty (Likert feedback), and (c) the median rank of the manipulated tests equal the default tests.
While the participants have an average error rate of 0.3 on default C-tests, the $T_i^{S,\mathrm{dec}}$ tests are significantly easier with an average error rate of 0.15 ($t = 7.49$, $p < 10^{-5}$) and the $T_i^{S,\mathrm{inc}}$ tests are significantly harder with an average error rate of 0.49 ($t = -7.83$, $p < 10^{-5}$), so we can safely reject the null hypothesis for error rates.

\begin{table}
	\centering\small
	\newcommand{\SIG}{${}^*$}
	\newcommand{\NSG}{\phantom{${}^*$}}
	\begin{tabular}{cccccc}
		\toprule
		& \multicolumn{2}{c}{easy (dec)} & default    & \multicolumn{2}{c}{hard (inc)} \\
		& $\SEL$ & $\SIZE$ & $\DEF$ & $\SEL$ & $\SIZE$ \\
		\midrule
		$T_1$ & -- & -- & .30 & -- & -- \\
		$T_2$ & .17\SIG & .11\SIG & .34 & .66\SIG & .44\SIG \\
		$T_3$ & .16\SIG & .10\SIG & .27 & .52\SIG & .43\SIG \\
		$T_4$ & .28\NSG & .09\SIG & .30 & .43\SIG & .45\SIG \\
		\midrule
		Average & .20\SIG & .10\SIG & .30 & .53\SIG & .44\SIG \\
		\bottomrule
	\end{tabular}
	\caption{Mean error rates $e(T)$ per text and strategy. Results marked with \SIG{} deviate significantly from $\DEF$}
	\label{tab:userstudy-error-rates-average}
\end{table}

\begin{figure}
	\centering
	\includegraphics[width=\linewidth,trim=0.2cm 0.4cm 0.65cm 0.35cm,clip]{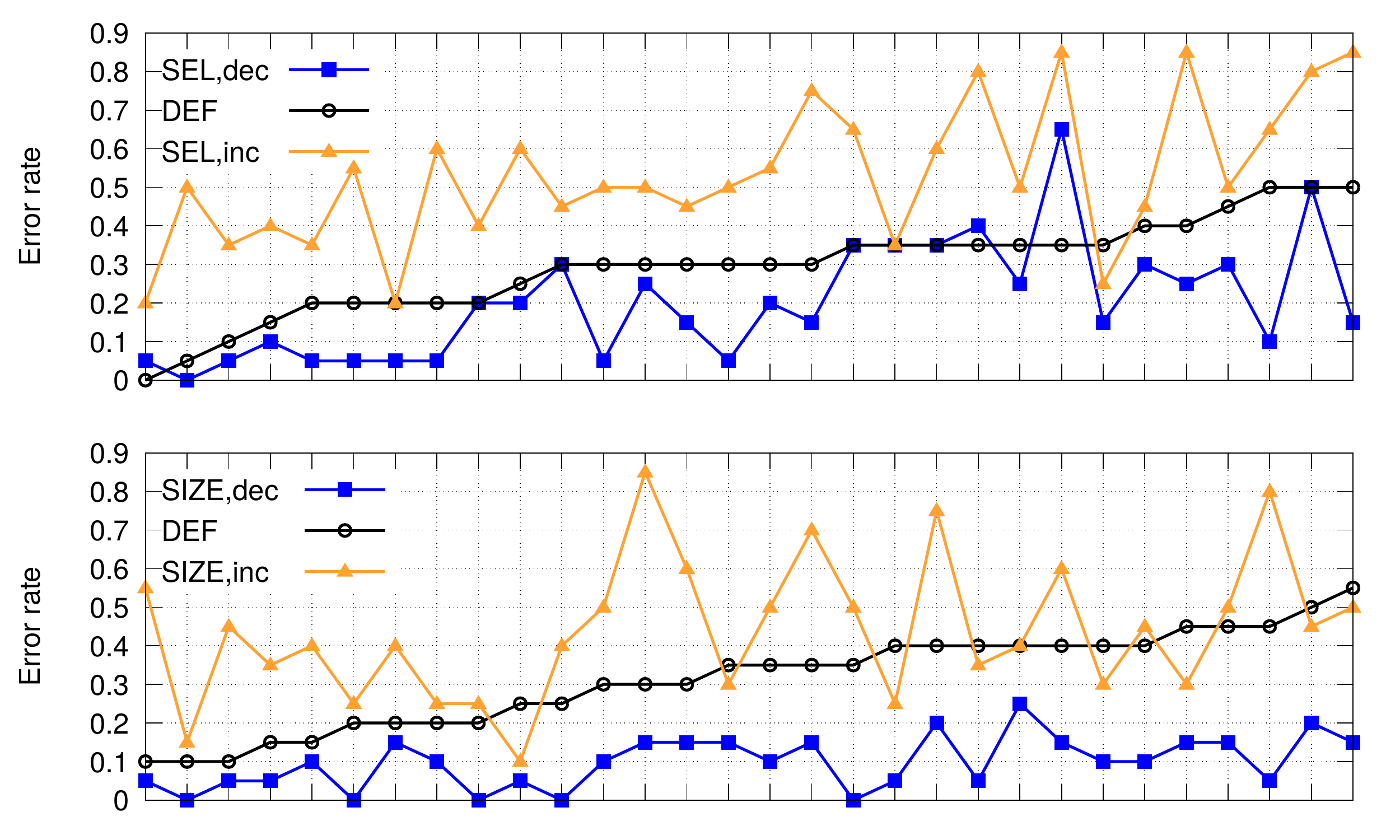}
	\caption{Error rates per participant and strategy}
	\label{fig:user-error-rates-def}
\end{figure}

Table~\ref{tab:userstudy-error-rates-average} shows the error rates per C-test and strategy. Both $\SEL$ and $\SIZE$ are overall able to significantly ($p < 0.025$) increase and decrease the test's difficulty over $\DEF$, and with the exception of $T_4^{\SEL,\mathrm{dec}}$, the effect is also statistically significant for all individual text and strategy pairs.
Figure~\ref{fig:user-error-rates-def} shows the 30 participants per strategy on the $x$-axis and their error rates in their second to fourth C-test on the $y$-axis.
C-tests, for which we increased the difficulty (${S,\mathrm{inc}}$), yield more errors than C-tests with decreased difficulty (${S,\mathrm{dec}}$) in all cases.
The easier tests also yield less errors than the test with the default scheme $\DEF$ in most cases.
While hard tests often have a much higher error rate than $\DEF$, we find some exceptions, in which the participant's error rate is close or even below the $\DEF$ error rate.

Regarding the perceived difficulty, we find that the participants judge the manipulated C-tests with lower $d(T)$ as easier on both the Likert scale ($z = 6.16$, $p < 10^{-5}$) and in the rankings  ($z = 6.59$, $p < 10^{-5}$) based on the Mann-Whitney-$U$ test.
The same is true for C-tests that have been manipulated to a higher difficulty level, which the participant judge harder ($z = -4.57$, $p < 10^{-5}$) and rank higher ($z = -3.86$, $p < 6 \cdot 10^{-5}$).
We therefore reject the null hypotheses for the Likert feedback and the rankings and conclude that both strategies can effectively manipulate a C-test's difficulty.

\paragraph{Manipulation quality.} 
We further investigate if the strategies yield different difficulty levels.
Therefore, we use two-tailed significance testing between $\SEL$ and $\SIZE$ for all three dimensions.
We find that $\SIZE$ yields significantly easier C-tests than $\SEL$ in terms of error rates ($p = 0.0014$) and Likert feedback ($p = 6 \cdot 10^{-5}$), and observe $p = 0.0394$ for the rankings.
For increasing the difficulty, we, however, do not find significant differences between the two strategies.
Since both strategies successfully modify the difficulty individually, this motivates research on combined strategies in the future.

\begin{table}
	\centering\small
	\newcommand{\SIG}{${}^*$}
	\newcommand{\NSG}{\phantom{${}^*$}}
	\begin{tabular}{lccccc}
		\toprule
		& \multicolumn{2}{c}{$\SEL$ } & $\DEF$    & \multicolumn{2}{c}{$\SIZE$} \\
		$\tau$ & .10 & .50 & -- & .10 & .50  \\
		\midrule
		\RMSE{}$(e,d)$ & .10 & .13  & .04 & .09 & .11 \\ 
		\RMSE{}$(e,\tau)$ & .12 & .10  & -- & .01 & .06 \\ 
		\bottomrule
	\end{tabular}
	\caption{\RMSE{} between the actual difficulty $e(T)$ and predicted difficulty $d(T)$ as well as target difficulty $\tau$.}
	\label{tab:userstudy-predicted-actual-average}
\end{table}

\begin{figure}
	\centering
	\includegraphics[width=0.50\textwidth]{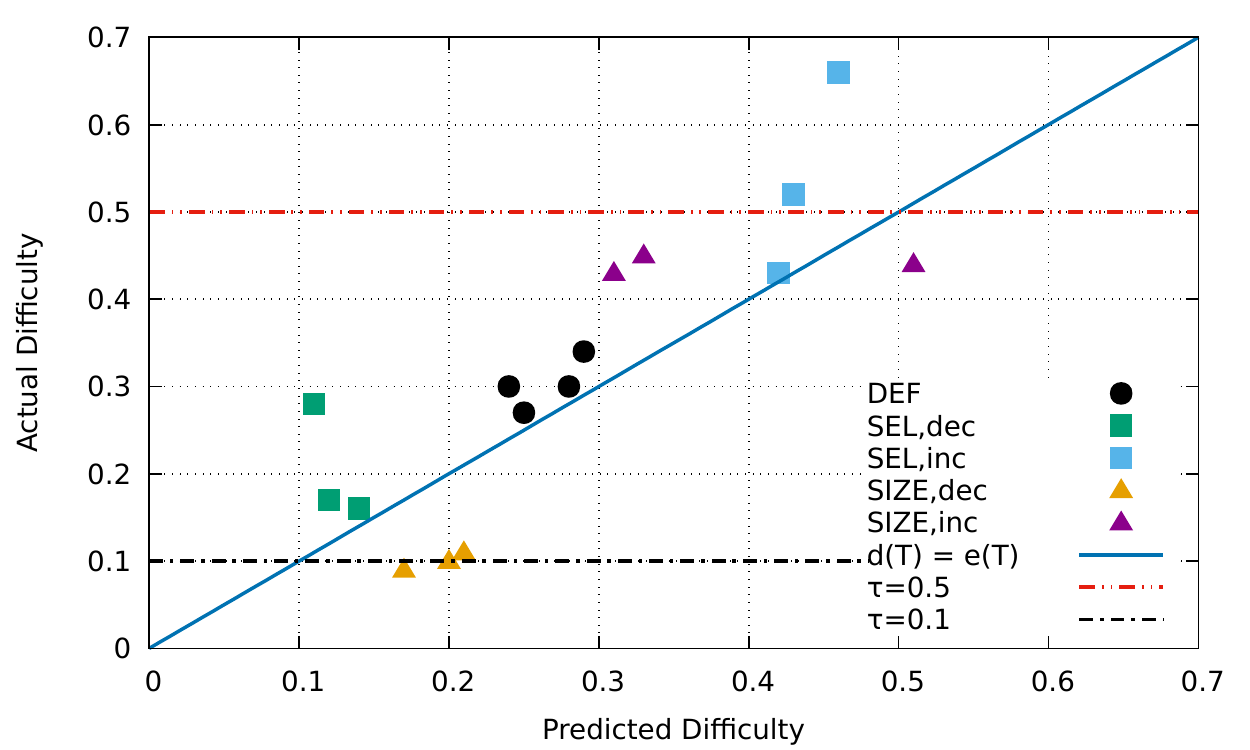}
	\caption{Predicted difficulties $d(T)$ vs the actual error rates $e(T)$.}
	\label{fig:predicted-vs-actual}
\end{figure}

We furthermore investigate how well our strategies perform in creating C-tests with the given target difficulty $\tau$.
Table~\ref{tab:userstudy-predicted-actual-average} shows the \RMSE{} for $e(T)$ and $d(T)$ as well as for $e(T)$ and $\tau$ for both strategies.
As expected, our difficulty prediction system works best for C-tests generated with $\DEF$ as they use the same scheme as C-tests in the training data.
Though slightly worse than for $\DEF$, we still find very low \RMSE{} scores for manipulated C-tests.
This is especially good when considering that the system's performance on our newly acquired dataset yields and \RMSE{} of $0.21$ (cf.\ section~\ref{sec:manip-eval}).
Computing the \RMSE{} with respect to our chosen target difficulties $\tau$ yields equally good results for $\SEL$ and exceptionally good results for $\SIZE$. 
Figure~\ref{fig:predicted-vs-actual} displays $d(T)$ in comparison to $e(T)$ for each individual text and strategy.
With the exception of $T_2^{\SEL,\mathrm{inc}}$ and $T_4^{\SEL,\mathrm{dec}}$, all 
predictions are close to the optimum (i.e., the diagonal) and also close to the desired target difficulty $\tau$.

In a more detailed analysis, we find two main sources of problems demanding further investigation:  
First, the difficulty prediction quality when deviating from $\DEF$ and second, the increasing ambiguity in harder C-tests. 
However, it underestimates the $d(T) = 0.11$ for $T_4^{\SEL,\mathrm{dec}}$ (the same text used in figure~\ref{fig:examples}), for which we found an actual error rate of $0.28$.
This is due to chains of four successive gaps, such as:
\medskip

\begin{tabular}{l|llll}
	gap $g$ & \texttt{i\_} & \texttt{wh\_\_} & \texttt{w\_} & \texttt{a\_\_} \\
	solution & \texttt{is} & \texttt{what} & \texttt{we} & \texttt{are} \\
	$d(g)$ & 0.17 & 0.22 & 0.23 & 0.19 \\
	$e(g)$ & 0.70 & 0.40 & 0.10 & 0.20\\
\end{tabular}
\medskip

\noindent
As the prediction system has been trained only on $\DEF$-generated C-tests, it underestimates $d(g)$ for cases with limited context.
It will be interesting for future work to focus on modeling gap interdependencies in C-tests deviating from $\DEF$.

Another issue we observe is that the gap size strategy might increase the ambiguity of the C-test.
In the standard scheme, there is in most cases only a single correct answer per gap.
In $T_2^{\SIZE, \mathrm{inc}}$, however, the $\SIZE$ strategy increased the gap of the word \textit{professional} to its maximal length yielding \texttt{p\_\_\_\_\_\_\_\_\_\_\_\_}.
One participant answered \textit{popularising} for this gap, which also fits the given context.
We carefully checked our dataset
for other ambiguity, but only found one additional case:
In $T_4$, instead of the word \textit{close}, 13 participants out of 30 used \textit{clear} as a modifier of \textit{correspondence}, which both produce meaningful contexts.
Given that this case is already ambiguous in the $\DEF$ scheme yielding the gap \texttt{cl\_\_\_}, we conclude that the issue is not severe, but that the difficulty prediction system should be improved to better capture ambiguous cases; for example, by introducing collocational features weighted by their distribution within a corpus into $\Delta_\mathrm{inc}$ and $\Delta_\mathrm{dec}$.

\section{Conclusion}
\label{sec:conclusion}

In this work, we proposed two novel strategies for automatically manipulating the difficulty of C-test exercises.
Our first strategy selects which words should be turned into a gap, and the second strategy learns to increase or decrease the size of the gaps. 
Both strategies automatically predict the difficulty of a test to make informed decisions.
To this end, we reproduced previous results, compared them to neural architectures, and tested them on a newly acquired dataset.
We evaluate our difficulty manipulation pipeline in a corpus-based study and with real users.
We show that both strategies can effectively manipulate the C-test difficulty, as both the participants' error rates and their perceived difficulty yield statistically significant effects.
Both strategies reach close to the desired difficulty level.

Our error analysis points out important directions for future work on detecting ambiguous gaps and modeling gap interdependencies for C-tests deviating from the default generation scheme.
An important observation is that manipulating the gaps' size and position does not only influence the C-test difficulty, but also addresses different competencies (e.g., requires more vocabulary knowledge or more grammatical knowledge).
Future manipulation strategies that take the competencies into account have the potential to train particular skills and to better control the competencies required for a placement test.
Another strand of research will be combining both strategies and deploying the manipulation strategies in a large scale testing platform  that allows the system to adapt to an individual learner over time.
A core advantage of our manipulation strategies is that we can work with any given text and thus provide C-tests that do not only have the desired difficulty, but also integrate the learner's interest or the current topic of a language course.

\section*{Acknowledgments}

This work has been supported by the Hessian research excellence program ``Landes-Offensive zur Entwicklung Wissenschaftlich-\"okonomischer Exzellenz'' (LOEWE) as part of the \emph{a!~-- automated language instruction} project under grant No.~521/17-03 and by the German Research Foundation as part of the Research Training Group ``Adaptive Preparation of Information from Heterogeneous Sources'' (AIPHES) under grant No.~GRK 1994/1.
We thank the anonymous reviewers for their detailed and helpful comments.
We furthermore thank the language center of the Technische Universit\"at Darmstadt for their cooperation and Dr.\ Lisa Beinborn for providing us with the code for our reproduction study.

\bibliographystyle{acl_natbib}
\bibliography{bibliography}

\begin{thebibliography}{31}
\expandafter\ifx\csname natexlab\endcsname\relax\def\natexlab#1{#1}\fi

\bibitem[{Beinborn et~al.(2014)Beinborn, Zesch, and Gurevych}]{Beinborn2014}
Lisa Beinborn, Torsten Zesch, and Iryna Gurevych. 2014.
\newblock \href {http://aclweb.org/anthology/Q14-1040} {{Predicting the
  Difficulty of Language Proficiency Tests}}.
\newblock \emph{Transactions of the Association for Computational Linguistics},
  2:517--529.

\bibitem[{Beinborn(2016)}]{Beinborn2016}
Lisa~Marina Beinborn. 2016.
\newblock \href {https://tuprints.ulb.tu-darmstadt.de/5647/} {\emph{Predicting
  and manipulating the difficulty of text-completion exercises for language
  learning}}.
\newblock Ph.D. thesis, Technische Universit\"at Darmstadt.

\bibitem[{Chandrasekar et~al.(1996)Chandrasekar, Doran, and
  Srinivas}]{Chandrasekar1996}
Raman Chandrasekar, Christine Doran, and Bangalore Srinivas. 1996.
\newblock \href {https://aclweb.org/anthology/C/C96/C96-2183.pdf} {Motivations
  and methods for text simplification}.
\newblock In \emph{Proceedings of the 16th International Conference on
  Computational Linguistics (COLING): Volume 2}, pages 1041--1044, Copenhagen,
  Denmark.

\bibitem[{Chapelle(1994)}]{Chapelle94}
C.~A. Chapelle. 1994.
\newblock \href {https://doi.org/10.1177/026765839401000203} {{Are C-tests
  valid measures for L2 vocabulary research?}}
\newblock \emph{Second Language Research}, 10(2):157--187.

\bibitem[{Chapelle and Abraham(1990)}]{Chapelle90}
Carol~A. Chapelle and Roberta~G. Abraham. 1990.
\newblock \href {https://doi.org/10.1177/026553229000700201} {{Cloze method:
  what difference does it make?}}
\newblock \emph{Language Testing}, 7(2):121--146.

\bibitem[{Collins-Thompson(2014)}]{Collins2014}
Kevyn Collins-Thompson. 2014.
\newblock \href {https://doi.org/10.1075/itl.165.2.01col} {Computational
  assessment of text readability: A survey of current and future research}.
\newblock \emph{International Journal of Applied Linguistics -- Special Issue
  on Recent Advances in Automatic Readability Assessment and Text
  Simplification}, 165(2):97--135.

\bibitem[{Dozat(2016)}]{Dozat2016}
Timothy Dozat. 2016.
\newblock \href {https://openreview.net/pdf?id=OM0jvwB8jIp57ZJjtNEZ}
  {Incorporating nesterov momentum into adam}.
\newblock In \emph{ICLR Workshop}.

\bibitem[{EC(2002)}]{EC02}
EC. 2002.
\newblock \href
  {http://ec.europa.eu/invest-in-research/pdf/download_en/barcelona_european_council.pdf}
  {{Presidency Conclusions. Barcelona European Council 15 and 16 March 2002}}.
\newblock Report SN 100/1/02 REV 1, Council of the European Union.

\bibitem[{Francis(1965)}]{Francis1965}
W.~Nelson Francis. 1965.
\newblock A standard corpus of edited present-day american english.
\newblock \emph{College English}, 26(4):267--273.

\bibitem[{Hancke et~al.(2012)Hancke, Vajjala, and Meurers}]{Hancke2012}
Julia Hancke, Sowmya Vajjala, and Detmar Meurers. 2012.
\newblock \href {http://aclweb.org/anthology/C12-1065} {Readability
  classification for german using lexical, syntactic, and morphological
  features}.
\newblock In \emph{Proceedings of the 24th International Conference on
  Computational Linguistics (COLING)}, pages 1063--1080, Mumbai, India.

\bibitem[{Hill and Simha(2016)}]{Hill2016}
Jennifer Hill and Rahul Simha. 2016.
\newblock \href {https://doi.org/10.18653/v1/W16-0503} {Automatic generation of
  context-based fill-in-the-blank exercises using co-occurrence likelihoods and
  google n-grams}.
\newblock In \emph{Proceedings of the 11th Workshop on Innovative Use of NLP
  for Building Educational Applications (BEA)}, pages 23--30, San Diego, CA,
  USA.

\bibitem[{Kamimoto(1993)}]{Kamimoto93}
Tadamitsu Kamimoto. 1993.
\newblock \href {https://doi.org/10.24539/llaj.30.0_47} {{Tailoring the Test to
  Fit the Students: Improvement of the C-Test through Classical Item
  Analysis}}.
\newblock \emph{Language Laboratory}, 30:47--61.

\bibitem[{Kilgarriff et~al.(2014)Kilgarriff, Charalabopoulou, Gavrilidou,
  Johannessen, Khalil, Kokkinakis, Lew, Sharoff, Vadlapudi, and
  Volodina}]{Kilgarriff2014}
Adam Kilgarriff, Frieda Charalabopoulou, Maria Gavrilidou, Janne~Bondi
  Johannessen, Saussan Khalil, Sofie~Johansson Kokkinakis, Robert Lew, Serge
  Sharoff, Ravikiran Vadlapudi, and Elena Volodina. 2014.
\newblock \href {https://doi.org/10.1007/s10579-013-9251-2} {Corpus-based
  vocabulary lists for language learners for nine languages}.
\newblock \emph{Language Resources and Evaluation}, 48(1):121--163.

\bibitem[{Klein-Braley and Raatz(1982)}]{Klein1982}
Christine Klein-Braley and Ulrich Raatz. 1982.
\newblock {Der C-Test: ein neuer Ansatz zur Messung allgemeiner
  Sprachbeherrschung}.
\newblock \emph{AKS-Rundbrief}, 4:23--37.

\bibitem[{Lahiri(2014)}]{Lahiri2014}
Shibamouli Lahiri. 2014.
\newblock \href {https://doi.org/10.3115/v1/E14-3011} {{Complexity of Word
  Collocation Networks: A Preliminary Structural Analysis}}.
\newblock In \emph{Proceedings of the Student Research Workshop at the 14th
  Conference of the European Chapter of the Association for Computational
  Linguistics}, pages 96--105, Gothenburg, Sweden.

\bibitem[{Laufer and Nation(1999)}]{Laufer99}
Batia Laufer and Paul Nation. 1999.
\newblock \href {https://doi.org/10.1177/026553229901600103} {{A
  vocabulary-size test of controlled productive ability}}.
\newblock \emph{Language Testing}, 16(1):33--51.

\bibitem[{Lee and Luo(2016)}]{Lee2016}
John Lee and Mengqi Luo. 2016.
\newblock \href {https://doi.org/10.18653/v1/P16-4020} {Personalized exercises
  for preposition learning}.
\newblock In \emph{Proceedings of the 54th Annual Meeting of the Association
  for Computational Linguistics (ACL): System Demonstrations}, pages 115--120,
  Berlin, Germany.

\bibitem[{Lewis et~al.(2004)Lewis, Yang, Rose, and Li}]{Lewis2004}
David~D. Lewis, Yiming Yang, Tony~G. Rose, and Fan Li. 2004.
\newblock \href {http://www.jmlr.org/papers/volume5/lewis04a/lewis04a.pdf}
  {{RCV1: A New Benchmark Collection for Text Categorization Research}}.
\newblock \emph{Journal of Machine Learning Research}, 5(Apr):361--397.

\bibitem[{Nisioi et~al.(2017)Nisioi, {\v{S}}tajner, Ponzetto, and
  Dinu}]{Nisioi2017}
Sergiu Nisioi, Sanja {\v{S}}tajner, Simone~Paolo Ponzetto, and Liviu~P. Dinu.
  2017.
\newblock \href {https://doi.org/10.18653/v1/P17-2014} {Exploring neural text
  simplification models}.
\newblock In \emph{Proceedings of the 55th Annual Meeting of the Association
  for Computational Linguistics (ACL): Short Papers}, volume~2, pages 85--91,
  Vancouver, Canada.

\bibitem[{Perez and Cuadros(2017)}]{Perez2017}
Naiara Perez and Montse Cuadros. 2017.
\newblock \href {http://aclweb.org/anthology/E17-3013} {Multilingual call
  framework for automatic language exercise generation from free text}.
\newblock In \emph{Proceedings of the 15th Conference of the European Chapter
  of the Association for Computational Linguistics (EACL): Software
  Demonstrations}, pages 49--52, Valencia, Spain.

\bibitem[{Pil{\'a}n et~al.(2014)Pil{\'a}n, Volodina, and Johansson}]{Pilan2014}
Ildik{\'o} Pil{\'a}n, Elena Volodina, and Richard Johansson. 2014.
\newblock \href {https://doi.org/10.3115/v1/W14-1821} {Rule-based and machine
  learning approaches for second language sentence-level readability}.
\newblock In \emph{Proceedings of the Ninth Workshop on Innovative Use of NLP
  for Building Educational Applications (BEA)}, pages 174--184, Baltimore, MD,
  USA.

\bibitem[{Reimers and Gurevych(2017)}]{Reimers2017}
Nils Reimers and Iryna Gurevych. 2017.
\newblock \href {https://aclweb.org/anthology/D17-1035} {{Reporting Score
  Distributions Makes a Difference: Performance Study of LSTM-networks for
  Sequence Tagging}}.
\newblock In \emph{Proceedings of the 2017 Conference on Empirical Methods in
  Natural Language Processing (EMNLP)}, pages 338--348, Copenhagen, Denmark.

\bibitem[{Sigott(1995)}]{Sigott1995}
G\"unther Sigott. 1995.
\newblock \href {https://www.jstor.org/stable/43023698} {{The C-Test: Some
  Factors of Difficulty}}.
\newblock \emph{AAA: Arbeiten aus Anglistik und Amerikanistik}, 20(1):43--53.

\bibitem[{Sigott(2006)}]{Sigott2006}
G{\"u}nther Sigott. 2006.
\newblock \href {https://www.peterlang.com/view/title/48699} {How fluid is the
  c-test construct?}
\newblock In \emph{Der C-Test: Theorie, Empirie, Anwendungen -- The C-Test:
  Theory, Empirical Research, Applications}, Language Testing and Evaluation,
  pages 139--146. Frankfurt am Main: Peter Lang.

\bibitem[{Spolsky(1969)}]{Spolsky69}
Bernard Spolsky. 1969.
\newblock \href {http://eric.ed.gov/?id=ED031702} {{Reduced Redundancy as a
  Language Testing Tool}}.
\newblock In G.E. Perren and J.L.M. Trim, editors, \emph{Applications of
  linguistics}, pages 383--390. Cambridge: Cambridge University Press.

\bibitem[{Taylor(1953)}]{Taylor1953}
Wilson~L. Taylor. 1953.
\newblock \href {https://doi.org/10.1177/107769905303000401} {{``Cloze
  Procedure'': A New Tool for Measuring Readability}}.
\newblock \emph{Journalism Bulletin}, 30(4):415--433.

\bibitem[{Vajjala and Meurers(2014)}]{Vajjala2014}
Sowmya Vajjala and Detmar Meurers. 2014.
\newblock \href {https://doi.org/10.3115/v1/E14-1031} {Assessing the relative
  reading level of sentence pairs for text simplification}.
\newblock In \emph{Proceedings of the 14th Conference of the European Chapter
  of the Association for Computational Linguistics (EACL)}, pages 288--297,
  Gothenburg, Sweden.

\bibitem[{Vapnik(1998)}]{Vapnik1998}
Vladimir~N. Vapnik. 1998.
\newblock \href {https://lccn.loc.gov/97037075} {\emph{Statistical Learning
  Theory}}.
\newblock New York: Wiley.

\bibitem[{Vygotsky(1978)}]{Vygotsky78}
Lev Vygotsky. 1978.
\newblock \emph{{Mind in society: The development of higher psychological
  processes}}.
\newblock Cambridge: Harvard University Press.

\bibitem[{Wojatzki et~al.(2016)Wojatzki, Melamud, and Zesch}]{Wojatzki2016}
Michael Wojatzki, Oren Melamud, and Torsten Zesch. 2016.
\newblock \href {https://doi.org/10.18653/v1/W16-0519} {Bundled gap filling: A
  new paradigm for unambiguous cloze exercises}.
\newblock In \emph{Proceedings of the 11th Workshop on Innovative Use of NLP
  for Building Educational Applications (BEA)}, pages 172--181, San Diego, CA,
  USA.

\bibitem[{Zesch and Melamud(2014)}]{Zesch2014}
Torsten Zesch and Oren Melamud. 2014.
\newblock \href {https://doi.org/10.3115/v1/W14-1817} {Automatic generation of
  challenging distractors using context-sensitive inference rules}.
\newblock In \emph{Proceedings of the Ninth Workshop on Innovative Use of NLP
  for Building Educational Applications (BEA)}, pages 143--148, Baltimore, MD,
  USA.

\end{thebibliography}

\appendix

\section{C-Test Difficulty Manipulation}
\paragraph{Feature description for $\Delta_\mathrm{inc}$ and $\Delta_\mathrm{dec}$.} We provide an extended feature description for the subset of features used for our relative difficulty prediction models $\Delta_\mathrm{inc}$ and $\Delta_\mathrm{dec}$. Features marked with * are also used by the absolute difficulty prediction model proposed by \citet{Beinborn2016}. For a gap $g = (i, \ell)$ in word $w_i$, we define:
\begin{itemize}
	\item the predicted absolute gap difficulty $d(g)$ for the initial C-test created with $\DEF$ 
	obtained from our reproduced difficulty prediction system, see line~3 of algorithm~2 (PS), 
	\item the word length $|w_i|$ (WL*),
	\item the new gap size $\ell \pm 1$ after modification (GL*), 
	\item the modified character $w_i[\ell]$ when increasing or decreasing the gap (CH), 
	\item a binary indicator if the gap is after a \textit{th} sound (RG*), and 
	\item the logarithmic difference of alternative solutions (LD*) capturing the change in the degree of ambiguity when increasing or decreasing $\ell$.
\end{itemize}

\paragraph{Feature ablation test.}
We conduct feature ablation tests to evaluate the impact of each feature on our relative difficulty prediction models $\Delta_\mathrm{inc}$ and $\Delta_\mathrm{dec}$.
Both models were evaluated on all gap size combinations for 120 random texts from the Brown corpus \cite{Francis1965} with a three-fold cross-validation. 
Table~\ref{tab:feature-ablation-gsist} shows the performance increase for each model after including each feature. 
\RMSE{} shows the deviation and $\rho$ the correlation of our relative difficulty prediction compared to the absolute difficulty prediction.
Although the increase in performance with RG is not substantial, we decided to include it as a meaningful feature which measures the impact for increasing or decreasing the gap size in words starting with \textit{th}.

\begin{table}
	\centering
	\begin{tabular}{l@{\hspace{.7cm}}c@{\quad}c@{\hspace{.7cm}}c@{\quad}c}
		\toprule
		& \multicolumn{2}{c}{$\Delta_\mathrm{inc}$} & \multicolumn{2}{c}{$\Delta_\mathrm{dec}$} \\
		Feature & \RMSE{} & $\rho$ & \RMSE{}  & $\rho$ \\
		\midrule
		PS & .088 & .521 & .213 & .271 \\
		+ WL & .072 & .712 & .183 & .570 \\
		+ GL & .066 & .771 & .162 & .687 \\
		+ CH & .069 & .735 & .157 & .707 \\
		+ RG & .069 & .736 & .157 & .707 \\
		+ LD & .061 & .805 & .131 & .806 \\
		\bottomrule
	\end{tabular}
	\caption{Feature ablation test for $\Delta_\mathrm{inc}$ and $\Delta_\mathrm{dec}$ compared to the full difficulty prediction system}
	\label{tab:feature-ablation-gsist}
\end{table}

\section{Neural Network Parameters}
\begin{figure}
	\small\centering
	\includegraphics[width=\linewidth]{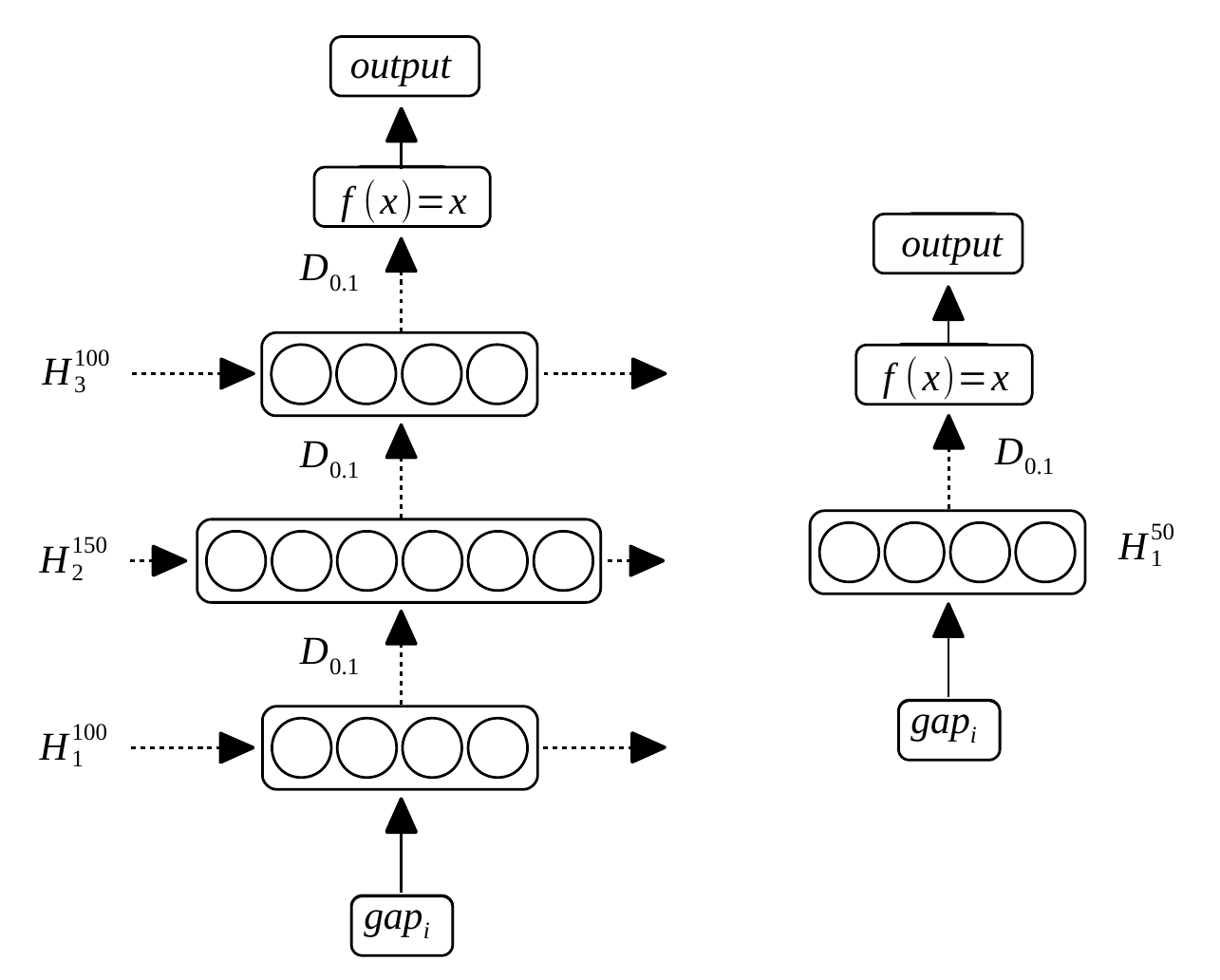}
	\caption{Final, tuned architectures of our BiLSTM (left) and MLP (right) models.}
	\label{fig:neural-models}
\end{figure}

Although obtaining state-of-the-art results in many tasks, the deep neural networks we evaluated during our preliminary experiments did perform worse than the SVM. 
We performed parameter tuning with 100 randomly initialized configurations for both, MLP and BiLSTM. 
We tune the following parameters:
\begin{itemize}
	\item Number of hidden layers $H_l \in [1, ..., 5]$
	\item Number of hidden units $H_l^u \in [50, ..., 200]$
	\item Dropout rate $D_x \in [0.1, ..., 0.5]$
\end{itemize}
We use Adam with Nesterov Momentum \cite{Dozat2016} as our optimizer and keep the batch size at $5$ for both models.
All models are trained for $200$ epochs with an early stopping after $10$ epochs with no improvement of the loss.
Figure \ref{fig:neural-models} shows the resulting architectures of both models after tuning. 
Since our goal is to output regression values, we use a linear activation function in the output layer. 

In preliminary experiments, we also tuned and evaluated BiLSTMs including soft attention, however, they performed even worse than the models without any attention.
Analyzing the results of the best performing attention based model showed that it had a strong bias towards predicting the mean value of the whole training set. 
Furthermore, similar to the other neural models, it showed a low error on the training set (low bias) and a rather high error on the development set (high variance), indicating a lack of training data. 

\section{Evaluation of the Manipulation System}

\begin{figure}
	\centering
	\includegraphics[width=.49\textwidth,trim=.1cm 0cm 0cm .25cm,clip]{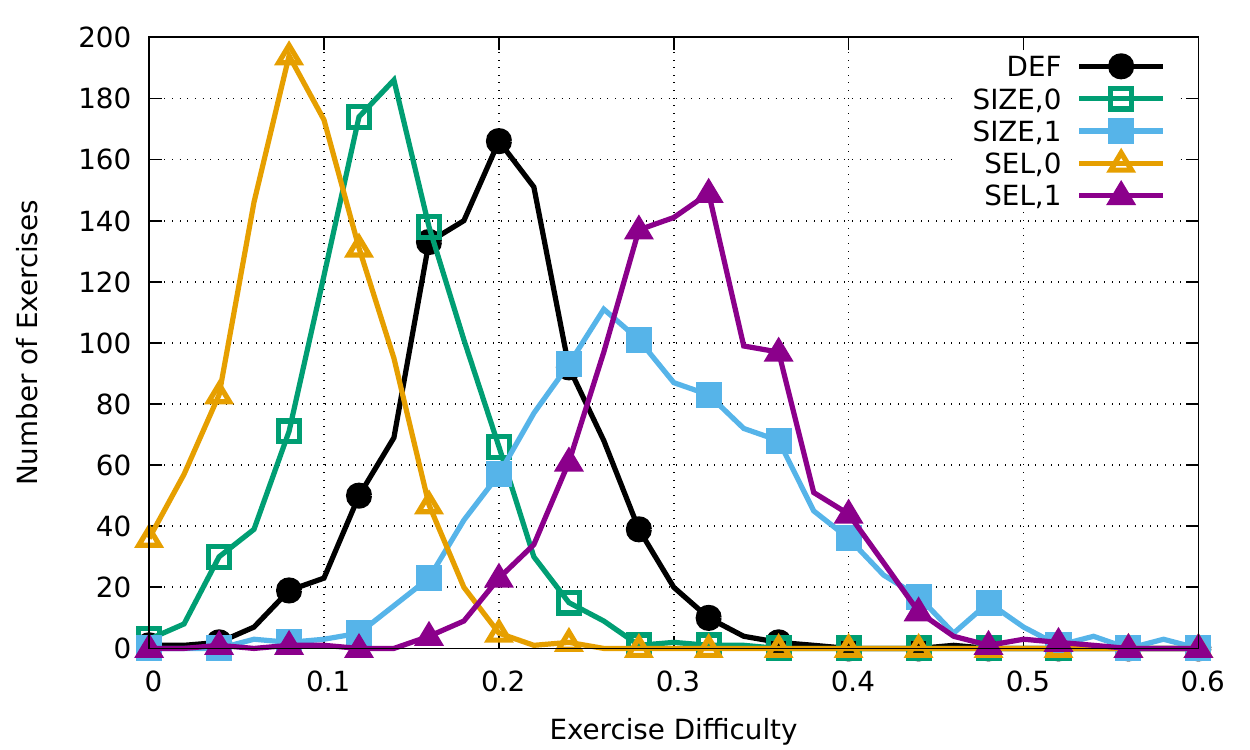}
	\caption{Difficulty distribution of exercises generated with $\DEF$, $\SEL$, and $\SIZE$ for extreme $\tau$ values on the Gutenberg corpus.}
	\label{fig:corpus-difficulties-gutenberg}
	\bigskip
	
	\centering
	\includegraphics[width=.49\textwidth,trim=.1cm 0cm 0cm .25cm,clip]{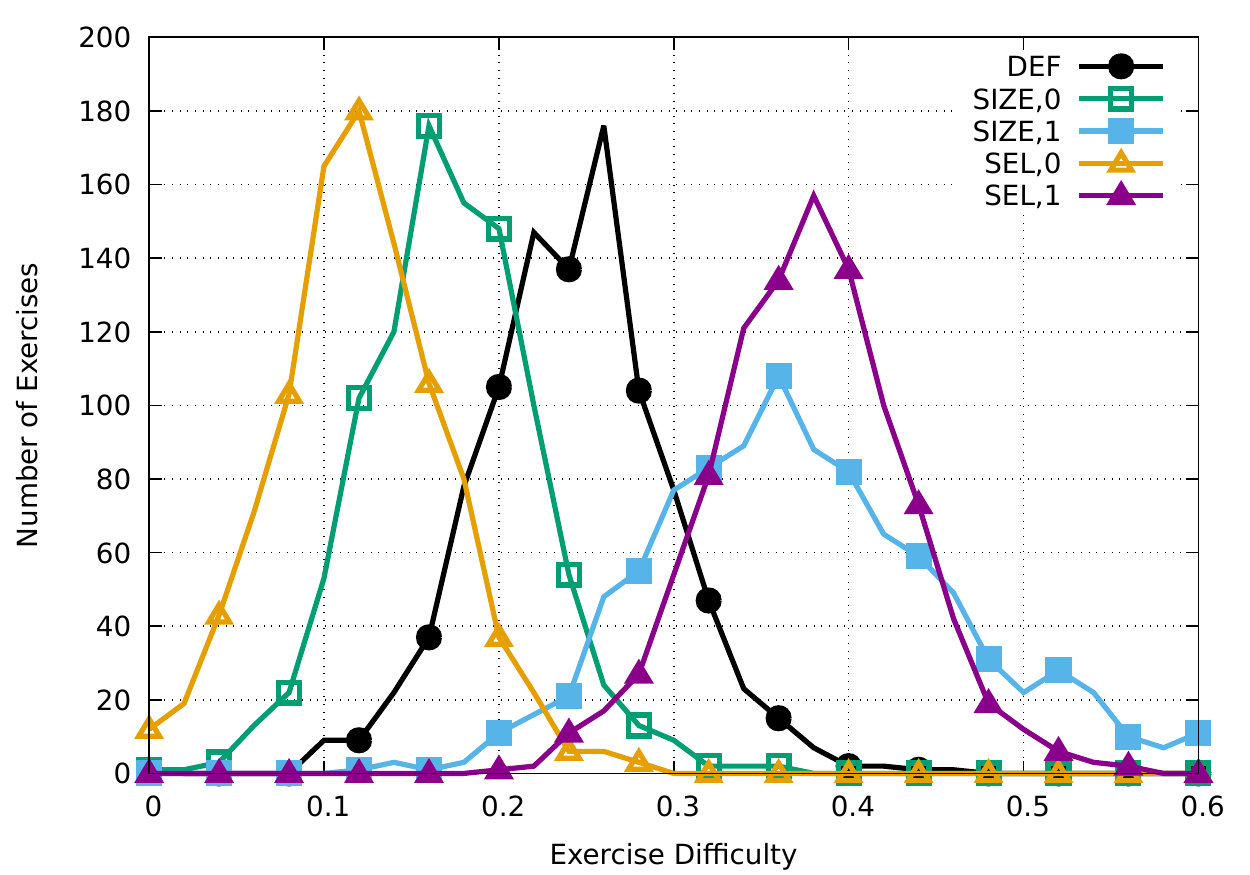}
	\caption{Difficulty distribution of exercises generated with $\DEF$, $\SEL$, and $\SIZE$ for extreme $\tau$ values on the Reuters corpus.}
	\label{fig:corpus-difficulties-reuters}
	\bigskip
	
	\centering
	\includegraphics[width=.49\textwidth,trim=.1cm 0cm 0cm .25cm,clip]{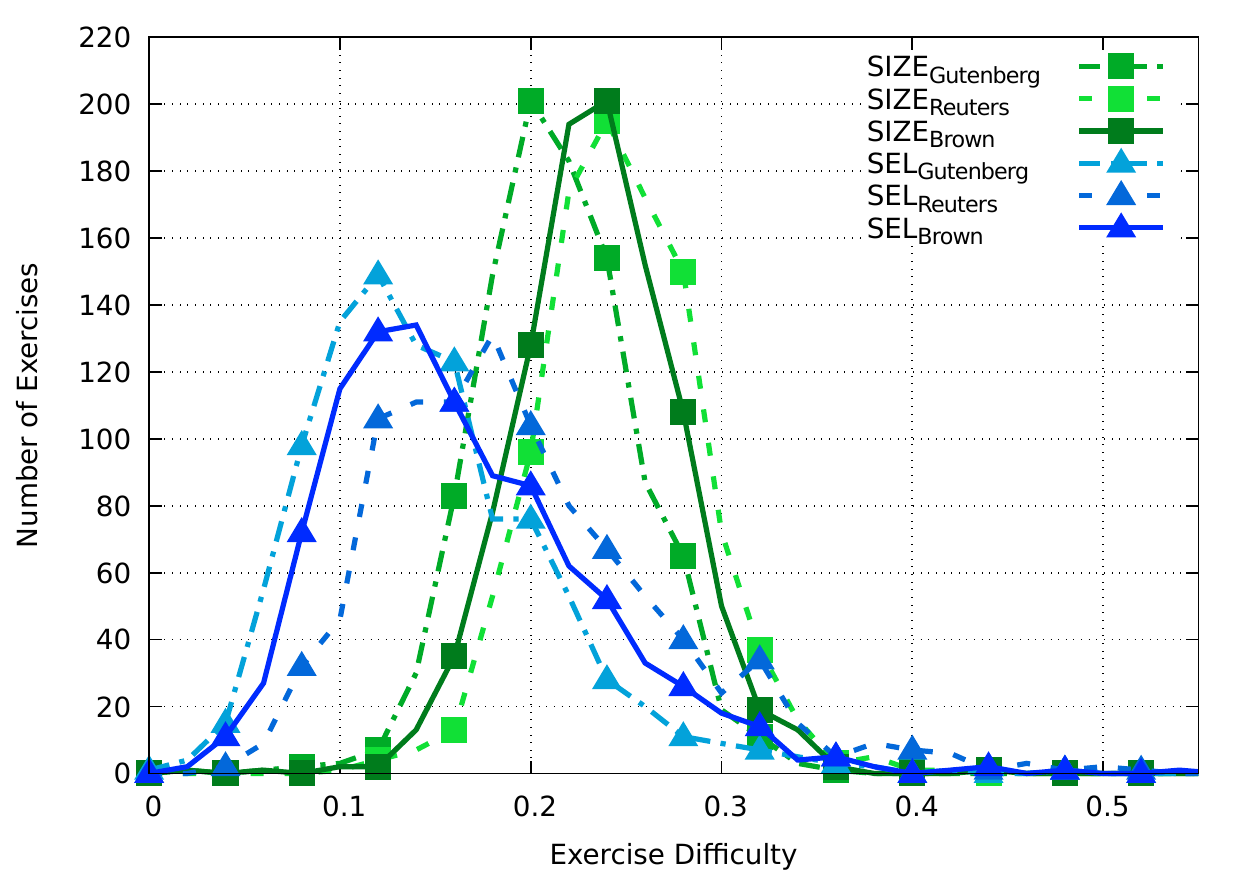}
	\caption{Error rate range ($\tau_\mathrm{max}-\tau_\mathrm{min}$) of exercises generated with $\SEL$ and $\SIZE$ for all three corpora.}
	\label{fig:base-difficulties-range}
\end{figure}

\paragraph{Results for additional corpora.} Figure~\ref{fig:corpus-difficulties-gutenberg} and figure~\ref{fig:corpus-difficulties-reuters} show our results on the Gutenberg \cite{Lahiri2014} and the Reuters \cite{Lewis2004} corpora. As already discussed in the main paper, we observe very similar distributions for $\DEF$, $\SEL$, and $\SIZE$ across both corpora matching our descriptions for the Brown \cite{Francis1965} corpus.

We further compute $\tau_\mathrm{max}-\tau_\mathrm{min}$ for $\SEL$ and $\SIZE$ for each text within a corpus and thus, measure the difficulty range both strategies are able to cover for a single text. 
As figure~\ref{fig:base-difficulties-range} shows, $\SEL$ achieves a larger difficulty range, whereas considerably more C-tests achieve higher difficulty levels when generated with $\SIZE$. 
We again observe very similar distributions throughout the three corpora.

\clearpage
\section{User-based Evaluation}

\paragraph{Questionnaire.}
At the begin of our study, our participants answered a questionnaire for a self-assessment of their English proficiency described in figure~\ref{fig:questionnaire}.
We partitioned our questionnaire into three sections asking about 1) our participants' \textit{English proficiency} (\textbf{Q1}, \textbf{Q2}), 2) their \textit{learning habits and goals} (\textbf{Q4}), and 3) \textit{other languages} they have been learning (\textbf{Q3}, \textbf{Q5}, \textbf{Q6}).

\begin{figure}[h]
	\newcommand{\BOX}[1]{\parbox[b]{#1}{\hrulefill}}
	\fbox{\parbox{\linewidth}{\raggedright
			\textbf{Q1}: Please estimate your current language proficiency in English \\
			\textbf{A1}: $\Circle$~\textit{Beginner (A1)} $\Circle$~\textit{Elementary (A2)} $\Circle$~\textit{Intermediate (B1)} $\Circle$~\textit{Upper Intermediate (B2)} $\Circle$~\textit{Advanced (C1)} $\Circle$~\textit{Proficiency (C2)} \medskip\\ 
			\textbf{Q2}: I studied English for about \BOX{.5cm} years. \medskip\\
			\textbf{Q3}: Do you participate in any language learning courses (for example, at your university, evening school,\dots)? If yes, than which ones? \\
			\textbf{A3}: \textit{$\Circle$ Yes, \BOX{1.5cm}.}  $\Circle$ \textit{No.} \medskip\\
			\textbf{Q4}: How often do you practice English? \\
			\textbf{A4}: $\Circle$ \textit{Never} $\Circle$ \textit{Monthly} $\Circle$ \textit{Weekly} $\Circle$ \textit{Daily} \medskip\\
			\textbf{Q5}: What is your native language?  \\
			\textbf{A5}: \BOX{4cm}\medskip\\
			\textbf{Q6}: Have you tried learning other languages before? If yes, than which ones?  \\
			\textbf{A6}: \textit{$\Circle$ Yes, \BOX{1.5cm}.}  $\Circle$ \textit{No.} 
		}}
		\caption{Self-assessment questionnaire.}\label{fig:questionnaire}
	\end{figure}
	
	\begin{figure}
		\small\centering
		\includegraphics[width=\linewidth]{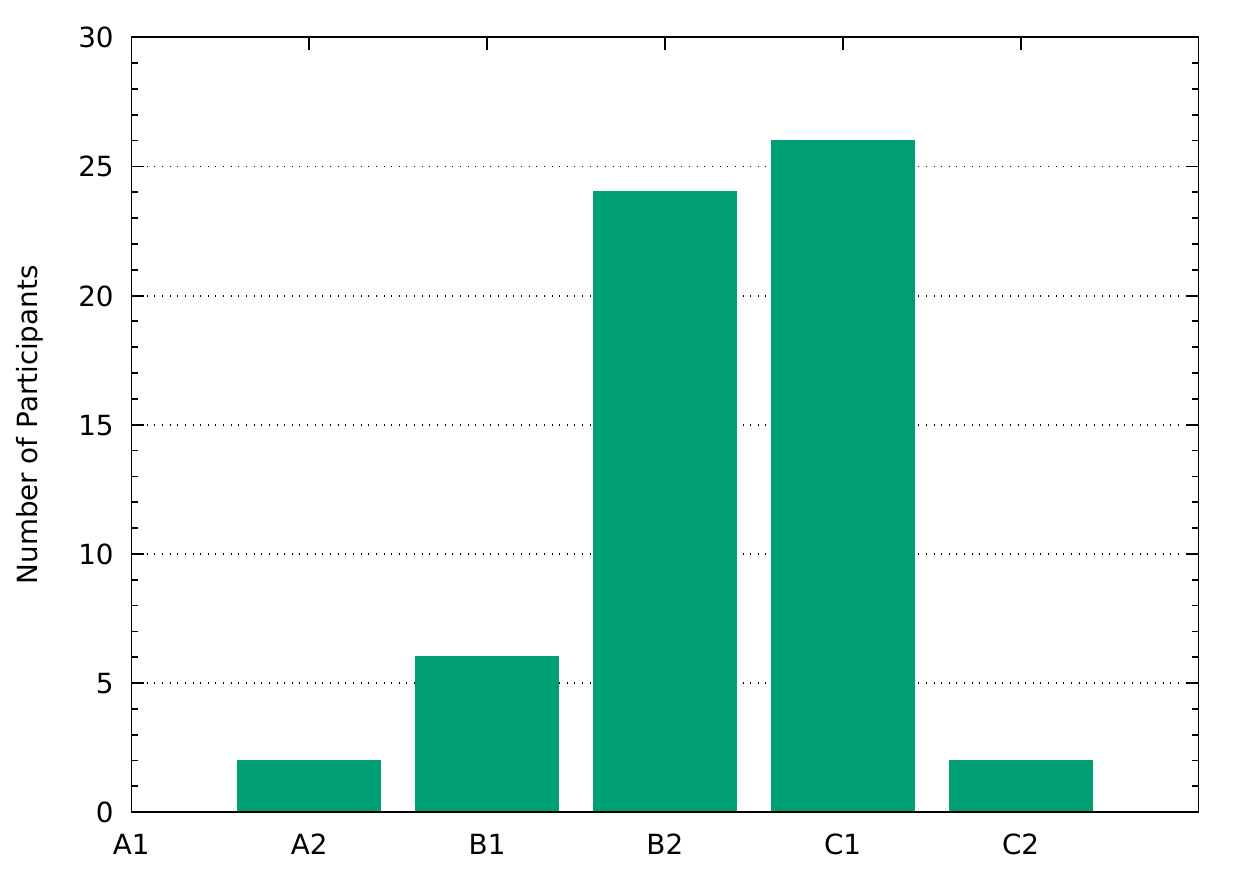}
		\caption{Our participants' CEFR level 
			self-assessment}
		\label{fig:self-assessment1}
		\bigskip
		
		\includegraphics[width=\linewidth]{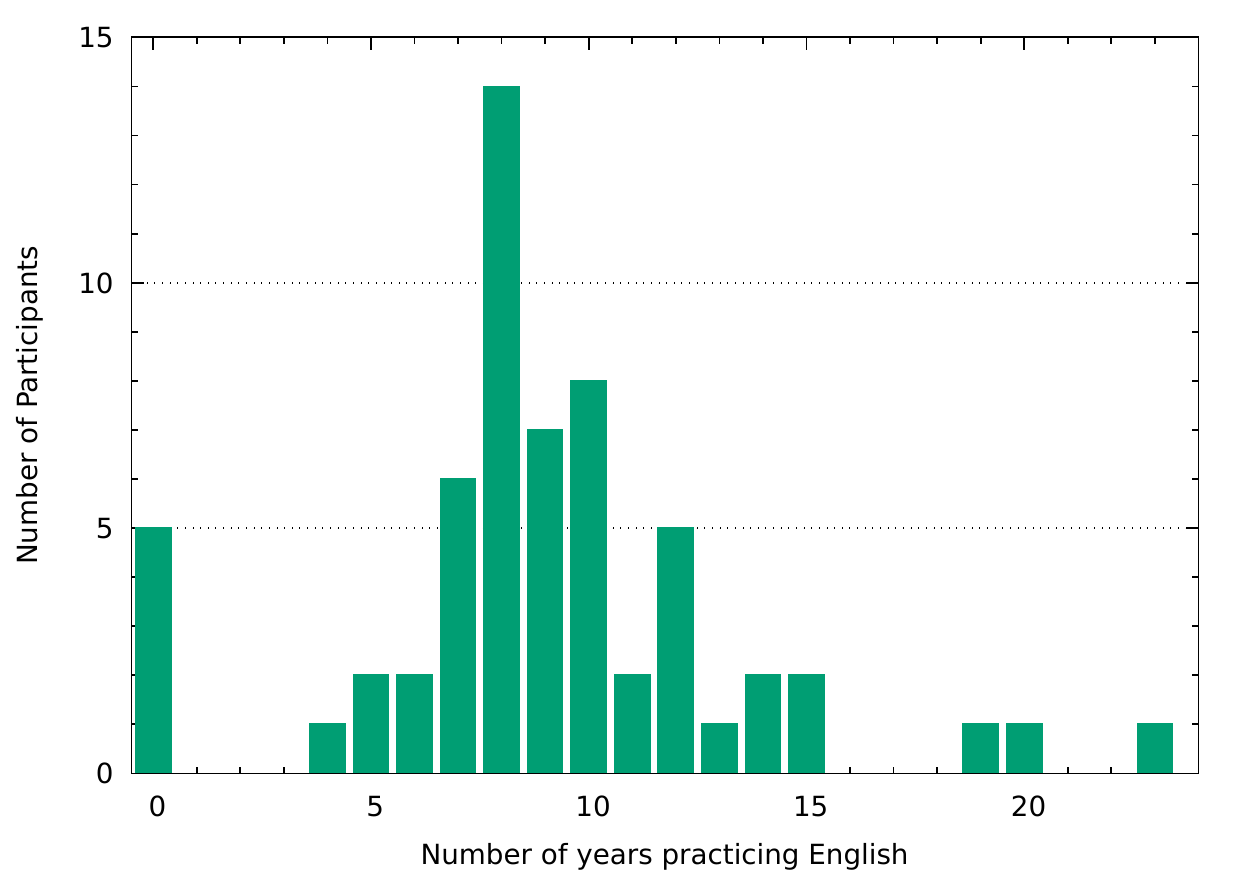}
		\caption{The number of years our participants have been practicing English}
		\label{fig:self-assessment2}
		\bigskip
		
		\includegraphics[width=\linewidth]{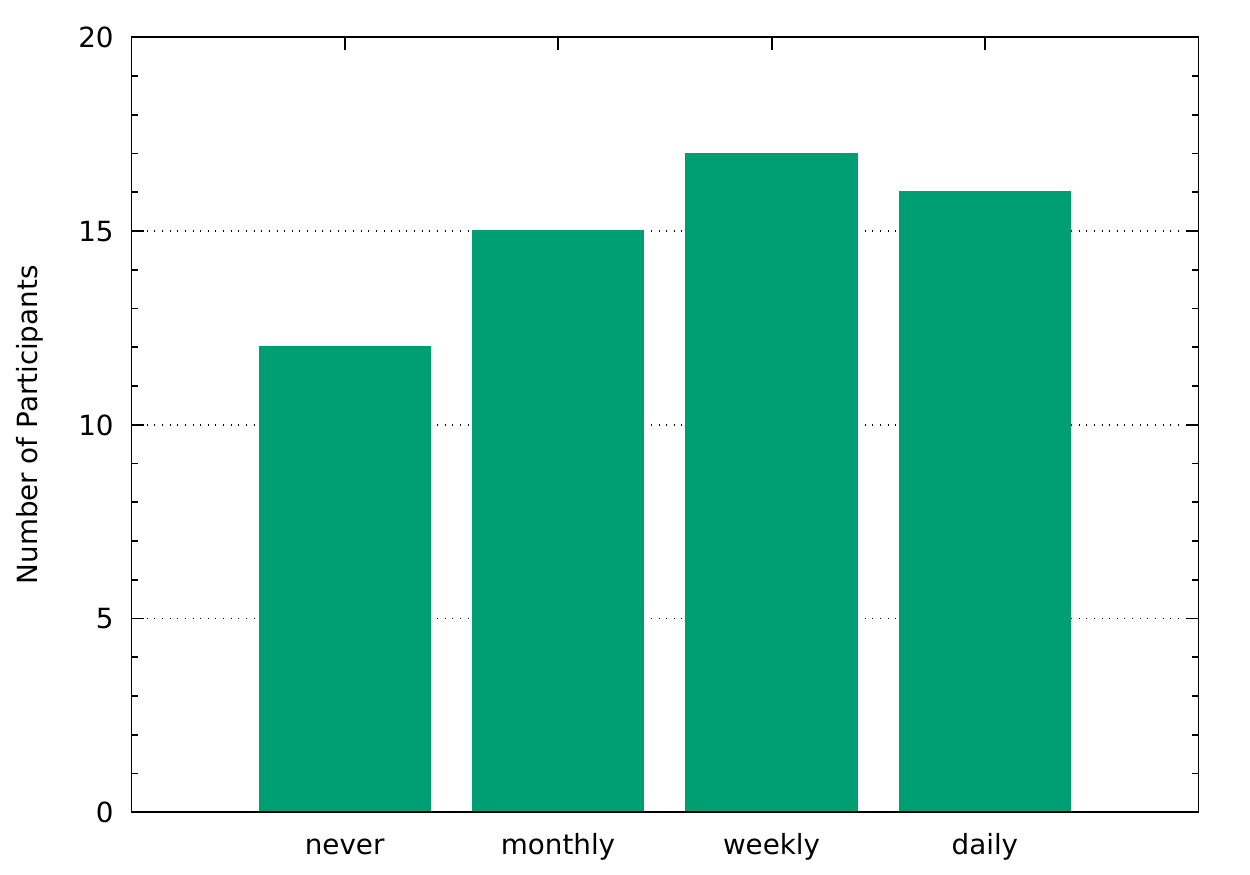}
		\caption{The frequency our participants have been practicing English}
		\label{fig:self-assessment3}
	\end{figure}

	\paragraph{Answers.} As described in the main paper, 17 participants are taking in language courses (\textbf{Q3}). Overall, 41 participants have tried to learn a second language (\textbf{Q6}). The exact answers can be found in the data we provide. Note, that not all participants provided the language which they attempted to learn since this was not mandatory. 
	Figure~\ref{fig:self-assessment1}--\ref{fig:self-assessment3} shows our participants' answers to \textbf{Q1}, \textbf{Q2}, and \textbf{Q4}. 
	As can be seen, none of our participants consider themselves at the \textit{Beginner (A1)} level. Furthermore, most of them are rather confident in their English proficiency and provide an estimate of either \textit{Upper Intermediate (B2)} or \textit{Advanced (C1)}.   
	
	\clearpage
	
	\paragraph{C-tests.}
	Figure~\ref{fig:ctestsDEF} shows the four texts $T_1$ to $T_4$ taken from the Brown corpus and the C-tests with the default gap scheme $\DEF$ we created from them for our user study.
	We have shortened each text to approximately 100 words and generated $n = 20$ gaps.
	In figure~\ref{fig:ctestsManip}, we provide the results of our manipulation strategies $\SEL$ and $\SIZE$ with decreased ($\tau = 0.1$) and increased ($\tau = 0.5$) difficulty.
	Note that, we only show sentences that contain gaps; the beginning and end of each text is the same as in figure~\ref{fig:ctestsDEF}.
	
	Table~\ref{tab:readability} reports readability scores for multiple common automated readability formulas.
	A Flesch reading ease score between 50--59 indicates \emph{fairly difficult}, 30--49 \emph{difficult}, and 0--29 \emph{very difficult}.
	A Gunning Fog score of 9.1 indicates \emph{fairly easy to read} and scores above 12 indicates \emph{hard to read}.
	The remaining readability scores corresponding to grade levels.
	
	\begin{figure}[t!]
		\begingroup
		\small
		\arrayrulecolor{gray}
		\newcolumntype{L}[1]{>{\ttfamily\small\raggedright\arraybackslash}m{#1}}
		\begin{tabular}{|*{1}{@{ }L{.45\textwidth}@{ }|}}
			\hline
			The study of the St. Louis area's economic prospects prepared for the Construction Industry Joint Conference confirms and reinforces both the findings of the Metropolitan St. Louis Survey of 1957 and the easily observed picture of the Missouri-Illinois countryside. St. Louis si\_\_ in t\_\_ center o\_ a relatively slow-growing a\_\_ in so\_\_ places stag\_\_\_\_ mid-continent region . Slac\_\_\_\_\_ regional dem\_\_\_ for St. Lo\_\_\_ goods a\_\_ services refl\_\_\_\_ the reg\_\_\_'s relative la\_\_ of purch\_\_\_\_\_ power. N\_\_ all St. Lo\_\_\_ industries, o\_ course, ha\_\_ a market ar\_\_ confined t\_ the immediate neighborhood. But for those which do, the slow growth of the area has a retarding effect on the metropolitan core.
			\\
			\hline
			\multicolumn{1}{c}{(a) C-test of $T_1$ with $\DEF$ gaps} \\[6pt]
			\hline
			Your invitation to write about Serge Prokofieff to honor his 70th Anniversary for the April issue of Sovietskaya Muzyka is accepted with pleasure, because I admire the music of Prokofieff; and with sober purpose, because the development of Prokofieff personifies, in many ways, the course of music in the Union of Soviet Socialist Republics. The Se\_\_\_ Prokofieff wh\_\_ we kn\_\_ in t\_\_ United Sta\_\_ of Ame\_\_\_\_ was g\_\_, witty, merc\_\_\_\_\_, full o\_ pranks a\_\_ bonheur -- a\_\_ very cap\_\_\_\_ as a profes\_\_\_\_\_\_ musician. Th\_\_\_ qualities ende\_\_\_\_ him t\_ both t\_\_ musicians a\_\_ the social-economic ha\_\_\_ monde wh\_\_\_ supported the concert world of the post-World War 1, era. Prokofieff's outlook as a composer-pianist-conductor in America was, indeed, brilliant.
			\\
			\hline
			\multicolumn{1}{c}{(b) C-test of $T_2$ with $\DEF$ gaps} \\[6pt]
			\hline
			The superb intellectual and spiritual vitality of William James was never more evident than in his letters. Here w\_\_ a man wi\_\_ an enor\_\_\_\_ gift f\_\_ living a\_ well a\_ thinking. T\_ both per\_\_\_\_ and id\_\_\_ he bro\_\_\_\_ the sa\_\_ delighted inte\_\_\_\_, the sa\_\_ open-minded relish f\_\_ what w\_\_ unique i\_ each, t\_\_ same discrim\_\_\_\_\_\_\_ sensibility a\_\_ quicksilver intell\_\_\_\_\_\_, the same gallantry of judgment. For this latest addition to the Great Letters Series, under the general editorship of Louis Kronenberger, Miss Hardwick has made a selection which admirably displays the variety of James's genius, not to mention the felicities of his style.
			\\
			\hline
			\multicolumn{1}{c}{(c) C-test of $T_3$ with $\DEF$ gaps} \\[6pt]
			Escalation unto death The nuclear war is already being fought, except that the bombs are not being dropped on enemy targets -- not yet. It i\_ being fou\_\_\_, moreover, i\_ fairly cl\_\_\_ correspondence wi\_\_ the predi\_\_\_\_\_\_ of t\_\_ soothsayers o\_ the th\_\_\_ factories. Th\_ predicted escal\_\_\_\_\_, and escal\_\_\_\_\_ is wh\_\_ we a\_\_ getting. T\_\_biggest nuc\_\_\_\_ device t\_\_ United Sta\_\_\_ has expl\_\_\_\_ measured so\_\_ 15 megatons, although our B-52s are said to be carrying two 20-megaton bombs apiece. Some time ago, however, Mr.\ Khrushchev decided that when bigger bombs were made, the Soviet Union would make them.
			\\
			\hline
			\multicolumn{1}{c}{(d) C-test of $T_4$ with $\DEF$ gaps} \\[-6pt]
		\end{tabular}
		\endgroup
		\caption{Standard C-tests of our user study}
		\label{fig:ctestsDEF}
	\end{figure}

	\begin{table}
		\small
		\begin{tabular}{l @{\hspace{.25cm}} cccc}
			\toprule
			Readability index & $T_1$ & $T_2$ & $T_3$ & $T_4$ \\
			\midrule
			Flesch reading ease & 56.1 & 24.8 & 32 & 55.6 \\
			Gunning Fog & 9.1 & 17.7 & 18.1 & 13.1 \\
			Flesch-Kincaid grade level & 8.2 & 17.3 & 15.2 & 9.6 \\
			Coleman-Liau index & 12 & 12 & 12 & 11  \\
			SMOG index & 8.1 & 15.5 & 13.5 & 10.1 \\
			Automated readability index & 7.9 & 17.4 & 15.5 & 9.7 \\
			Linsear Write formula & 6.5 & 22.3 & 18.4 & 11.2 \\
			\bottomrule
		\end{tabular}
		\caption{Automated readability analysis of the four texts used for our C-tests. Scores are based on the online tool at \url{http://www.readabilityformulas.com}.}
		\label{tab:readability}
	\end{table}

	\FloatBarrier

	\begin{figure*}
		\begingroup
		\small
		\arrayrulecolor{gray}
		\newcolumntype{L}[1]{>{\ttfamily\small\raggedright\arraybackslash}m{#1}}
		\begin{tabular}{|*{2}{@{ }L{.49\textwidth}@{ }|}}
			\hline
			\dots The Serg\_ Prokofieff who\_ we kne\_ in t\_\_ United State\_ of Americ\_ was ga\_, witty, mercuria\_, full o\_ pranks an\_ bonheur -- an\_ very capabl\_ as a professiona\_ musician. Thes\_ qualities endeare\_ him t\_ both t\_\_ musicians an\_ the social-economic haut\_ monde whic\_ supported\dots
			&
			\dots The S\_\_\_\_ Prokofieff wh\_\_ we kn\_\_ in t\_\_ United S\_\_\_\_\_ of A\_\_\_\_\_\_ was ga\_, witty, mercu\_\_\_\_, full o\_ pranks a\_\_ bonheur -- a\_\_ very cap\_\_\_\_ as a \linebreak p\_\_\_\_\_\_\_\_\_\_\_ musician. T\_\_ qualities end\_\_\_\_\_ him t\_ both t\_\_ musicians a\_\_ the social-economic h\_\_\_\_ monde wh\_\_\_ supported\dots
			\\ \hline
			\multicolumn{1}{c}{(a) C-test of $T_2$ manipulated with $\SIZE$ for $\tau = 0.1$} & 
			\multicolumn{1}{c}{(b) C-test of $T_2$ manipulated with $\SIZE$ for $\tau = 0.5$}
			\\[6pt]
			\hline
			\dots T\_\_ Serge Proko\_\_\_\_\_ whom w\_ kn\_\_ i\_ t\_\_ Uni\_\_\_ Sta\_\_\_ o\_ Ame\_\_\_ w\_\_ gay, witty, mercurial, fu\_\_ o\_ pranks and bonheur -- a\_\_ ve\_\_ capable a\_\_ a professional musician. These qualities endeared h\_\_ t\_ both t\_\_ musicians a\_\_ the social-economic haute monde which supported\dots
			&
			\dots The Se\_\_\_ Prokofieff wh\_\_ we kn\_\_ in the United States of America was g\_\_, wi\_\_\_, merc\_\_\_\_\_, full of pra\_\_\_ a\_\_ bon\_\_\_\_ -- and very cap\_\_\_\_ as a \linebreak profes\_\_\_\_\_\_ musi\_\_\_\_. Th\_\_\_ qual\_\_\_\_\_ ende\_\_\_\_ h\_\_ to bo\_\_ the musi\_\_\_\_ and the social-economic ha\_\_\_ \linebreak mo\_\_\_ which supported\dots
			\\ \hline
			\multicolumn{1}{c}{(c) C-test of $T_2$ manipulated with $\SEL$ for $\tau = 0.1$} & 
			\multicolumn{1}{c}{(d) C-test of $T_2$ manipulated with $\SEL$ for $\tau = 0.5$}
			\\[6pt]
			\hline
			\dots Here wa\_ a man wit\_ an enormou\_ gift fo\_ living a\_ well a\_ thinking. T\_ both person\_ and idea\_ he brough\_ the sa\_\_ delighted interes\_, the sa\_\_ open-minded relish fo\_ what wa\_ unique i\_ each, t\_\_ same discriminatin\_ sensibility an\_ quicksilver intelligenc\_, the same gallantry of judgment\dots
			&
			\dots Here w\_\_ a man w\_\_\_ an e\_\_\_\_\_\_\_ gift f\_\_ living a\_ well a\_ thinking. T\_ both per\_\_\_\_ and id\_\_\_ he \linebreak bro\_\_\_\_ the s\_\_\_ delighted inte\_\_\_\_, the s\_\_\_ open-minded relish f\_\_ what w\_\_ unique i\_ each, t\_\_ same d\_\_\_\_\_\_\_\_\_\_\_\_ sensibility a\_\_ quicksilver \linebreak i\_\_\_\_\_\_\_\_\_\_\_, the same gallantry of judgment\dots
			\\ \hline
			\multicolumn{1}{c}{(e) C-test of $T_3$ manipulated with $\SIZE$ for $\tau = 0.1$} & 
			\multicolumn{1}{c}{(f) C-test of $T_3$ manipulated with $\SIZE$ for $\tau = 0.5$}
			\\[6pt]
			\hline
			\dots Here w\_\_ a m\_\_ wi\_\_ a\_ enormous gift f\_\_ liv\_\_\_ a\_ we\_\_ a\_ thinking. T\_ both persons and ideas h\_ \linebreak bro\_\_\_\_ t\_\_ sa\_\_ delighted interest, t\_\_ sa\_\_ open-minded relish f\_\_ what w\_\_ unique i\_ each, t\_\_ same discriminating sensibility and quicksilver intelligence, the same gallantry of judgment\dots
			&
			\dots He\_\_ was a m\_\_ with an enor\_\_\_\_ gi\_\_ for living as well as thin\_\_\_\_. T\_ bo\_\_ per\_\_\_\_ a\_\_ id\_\_\_ he \linebreak brought the same deli\_\_\_\_\_ inte\_\_\_\_, the same open-minded rel\_\_\_ for wh\_\_ was uni\_\_\_ in ea\_\_, the same discrim\_\_\_\_\_\_\_ sensi\_\_\_\_\_\_ a\_\_ quick\_\_\_\_\_\_ intelligence, the same gallantry of judgment\dots
			\\ \hline
			\multicolumn{1}{c}{(g) C-test of $T_3$ manipulated with $\SEL$ for $\tau = 0.1$} & 
			\multicolumn{1}{c}{(h) C-test of $T_3$ manipulated with $\SEL$ for $\tau = 0.5$}
			\\[6pt]
			\hline
			\dots It i\_ being fough\_, moreover, i\_ fairly clos\_ correspondence wit\_ the prediction\_ of t\_\_ soothsayers o\_ the thin\_ factories. The\_ predicted escalatio\_, and escalatio\_ is wha\_ we ar\_ getting. T\_\_ biggest nuclea\_ device t\_\_ United State\_ has explode\_ measured som\_ 15 megatons\dots
			&
			\dots It i\_ being fou\_\_\_, moreover, i\_ fairly c\_\_\_\_ correspondence w\_\_\_ the p\_\_\_\_\_\_\_\_\_\_ of t\_\_ soothsayers o\_ the th\_\_\_ factories. T\_\_\_ predicted es\_\_\_\_\_\_\_\_\_, and es\_\_\_\_\_\_\_\_\_ is wh\_\_ we a\_\_ getting. T\_\_ biggest nu\_\_\_\_\_ device t\_\_ United Sta\_\_\_ has expl\_\_\_\_ measured s\_\_\_ \linebreak 15 megatons\dots
			\\ \hline    
			\multicolumn{1}{c}{(i) C-test of $T_4$ manipulated with $\SIZE$ for $\tau = 0.1$} & 
			\multicolumn{1}{c}{(j) C-test of $T_4$ manipulated with $\SIZE$ for $\tau = 0.5$}
			\\[6pt]
			\hline
			\dots I\_ i\_ be\_\_\_ fou\_\_\_, moreover, i\_ fairly close correspondence wi\_\_ t\_\_ predictions o\_ t\_\_ soothsayers o\_ t\_\_ think factories. They predicted escalation, a\_\_ escalation i\_ wh\_\_ w\_ a\_\_ getting. T\_\_ big\_\_\_\_ nuclear device t\_\_ Uni\_\_\_ States has exploded measured some 15 megatons\dots
			&
			\dots It is being fought, more\_\_\_\_, in fai\_\_\_ cl\_\_\_ corresp\_\_\_\_\_\_\_ with the predi\_\_\_\_\_\_ of the sooth\_\_\_\_\_\_ \linebreak of the th\_\_\_ fact\_\_\_\_\_. Th\_\_ pred\_\_\_\_\_ escal\_\_\_\_\_, and escal\_\_\_\_\_ is what w\_ are get\_\_\_\_. The big\_\_\_\_ nuc\_\_\_\_ dev\_\_\_ the United States h\_\_ expl\_\_\_\_ meas\_\_\_\_ some \linebreak 15 megatons\dots
			\\
			\hline
			\multicolumn{1}{c}{(k) C-test of $T_4$ manipulated with $\SEL$ for $\tau = 0.1$} & 
			\multicolumn{1}{c}{(l) C-test of $T_4$ manipulated with $\SEL$ for $\tau = 0.5$}
			\\
		\end{tabular}
		\endgroup
		\caption{Manipulated C-tests of our user study}
		\label{fig:ctestsManip}
	\end{figure*}
	
	\FloatBarrier


\end{document}